\documentclass[letterpaper, 10 pt, conference]{ieeeconf}  

\IEEEoverridecommandlockouts                              

\usepackage{times}
\usepackage{epsfig}
\usepackage{graphicx}
\usepackage{amsmath}
\usepackage{amssymb}
\usepackage{cite}
\usepackage{amsmath,amssymb,amsfonts}
\usepackage{textcomp}
\usepackage{subfigure}
\usepackage[linesnumbered,boxed,ruled,commentsnumbered]{algorithm2e}
\usepackage{algpseudocode}  
\usepackage{amsmath}  
\usepackage{multirow}
\UseRawInputEncoding

\usepackage{url}

\title{\LARGE \bf
Outlier-Robust Geometric Perception: A Novel Thresholding-Based Estimator with Intra-Class Variance Maximization
}

\author{Lei Sun$^{1}$
\thanks{*This paper has been accepted to IROS2024}
\thanks{$^{1}$Lei Sun is with School of Engineering and Applied Science, University of Pennsylvania, Philadelphia, USA \,(e-mail: leisunjames@126.com)
}%
}

\begin{document}

\maketitle
\thispagestyle{empty}
\pagestyle{empty}

\begin{abstract}

Geometric perception problems are fundamental tasks in robotics and computer vision. In real-world applications, they often encounter the inevitable issue of outliers, preventing traditional algorithms from making correct estimates. In this paper, we present a novel general-purpose robust estimator TIVM (Thresholding with Intra-class Variance Maximization) that can collaborate with standard non-minimal solvers to efficiently reject outliers for geometric perception problems. First, we introduce the technique of intra-class variance maximization to design a dynamic 2-group thresholding method on the measurement residuals, aiming to distinctively separate inliers from outliers. Then, we develop an iterative framework that robustly optimizes the model by approaching the pure-inlier group using a multi-layered dynamic thresholding strategy as subroutine, in which a self-adaptive mechanism for layer-number tuning is further employed to minimize the user-defined parameters. We validate the proposed estimator on 3 classic geometric perception problems: rotation averaging, point cloud registration and category-level perception, and experiments show that it is robust against 70--90\% of outliers and can converge typically in only 3--15 iterations, much faster than state-of-the-art robust solvers such as RANSAC, GNC and ADAPT. Furthermore, another highlight is that: our estimator can retain approximately the same level of robustness even when the inlier-noise statistics of the problem are fully unknown. 

\end{abstract}

\section{INTRODUCTION}

Geometric spatial perception problems, which aim to estimate the geometric models regarding the state of a robot or a sensor (e.g. camera) and its spatial relationship to the environment~\cite{tzoumas2019outlier}, are the building blocks in robotics and computer vision, having found broad applications in motion estimation~\cite{zhang2015visual}, object/camera localization~\cite{zeng2017multi,wong2017segicp}, 3D reconstruction~\cite{blais1995registering,choi2015robust}, autonomous navigation and SLAM~\cite{cadena2016past}. 

Unfortunately, in realistic applications, geometric perception problems are often not only affected by noisy data, but also corrupted by the {outliers}, which are spurious measurements typically caused by wrong data association/matches, extreme noise or device malfunction, making traditional estimation methods fail to yield correct results.

\begin{figure}[ht]
\centering
\begin{tabular}{c}
\begin{minipage}[t]{0.93\linewidth}
\centering
\includegraphics[width=1\linewidth]{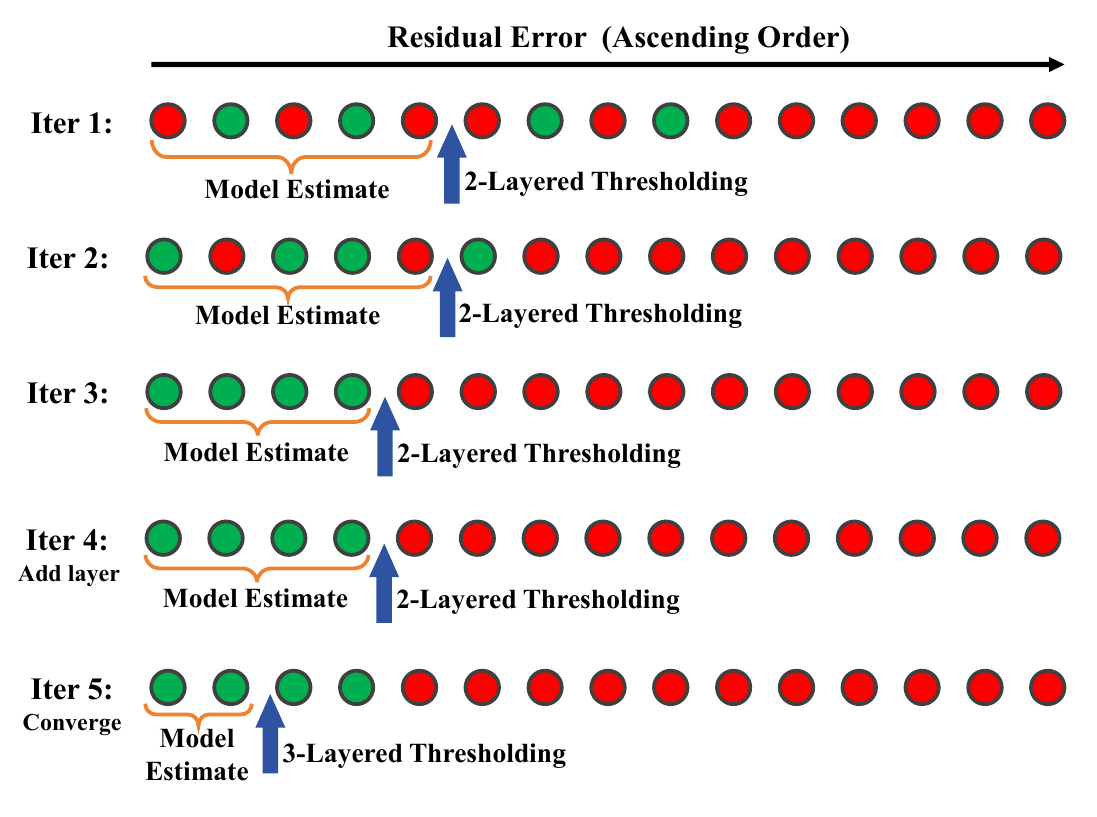}
\end{minipage}
\end{tabular}
\vspace{-3mm}
\caption{Intuitive illustration on the thresholding framework of the proposed TIVM estimator in Algorithm~\ref{algo-TIVM*}. }
\label{illus-demo}
\vspace{-1mm}
\end{figure}

In terms of handling outliers, RANSAC~\cite{fischler1981random} is the most classic robust heuristic widely applied in many perception problems to remove outliers by incorporating the {minimal} solver into a hypothesize-and-check paradigm. Though some variants (e.g. ~\cite{chum2003locally,chum2005matching,sun2022ransic}) are proposed to improve the performance, RANSAC-based estimators usually suffer from exponentially increasing computational cost w.r.t. the outlier ratio as well as the minimal subset size, inefficient for use in many real-world tasks. Moreover, minimal solvers may not fit with all the problems, as will be discussed in Section~\ref{sec-minimal}.

More recently, global non-minimal robust estimators such as the GNC framework~\cite{zhou2016fast,yang2020graduated,peng2023convergence} and ADAPT~\cite{tzoumas2019outlier} are receiving increasing attention, since they can reject outliers directly in conjunction with standard {non-minimal} solvers that have been rapidly developed these years for multiple perception problems (e.g.~\cite{yang2020perfect} for 2D-3D shape reconstruction,~\cite{shi2021optimal} for category-level perception, and~\cite{rosen2019se} for pose graph optimization) without the need of initial guess. However, although they can reject 70--80\% outliers, they are generally too slow to converge, requiring at least dozens of, or even over a hundred, iterations to attain convergence. Since most non-minimal solvers are not as fast as the minimal ones, using these non-minimal robust estimators could result in long computational time. 

In addition, another challenge in the geometric perception problems is that: a user-defined inlier threshold is needed to differentiate inliers from outliers during robust estimation, but determining such inlier threshold can be difficult since it usually requires users to manually collect sufficient data to estimate the noise statistics or complete an arduous work of parameter tuning. Besides, the inlier-noise statistics may vary during actual tasks (e.g. due to change of weather, long-time usage), so the original calibration could become inaccurate.

\textbf{Contributions.} To address these limitations, this paper proposes a novel general-purpose non-minimal robust estimation method TIVM (Thresholding with Intra-class Variance Maximization) for geometric perception, as intuitively illustrated in Fig.~\ref{illus-demo}. First, we present a dynamic residual thresholding method based on intra-class variance maximization seeking to flexibly classify the measurements into a low-magnitude group (approaching inliers) and a high-magnitude group (approaching outliers). Second, an iterative model-optimizing framework is developed upon a multi-layered dynamic thresholding method to guarantee high robustness. Moreover, a self-adaptive strategy to tune the layer number is invented to enable automatic algorithm operation without introducing additional user-defined parameters. More importantly, the resulting estimator TIVM could achieve robust estimation even without any prior knowledge of the inlier-noise statistics. We test TIVM on the rotation averaging, point cloud registration, and category-level perception problems. TIVM demonstrates 70--90\% robustness against outliers no matter the noise-statistics are known or unknown, and it could converge in merely 3--15 iterations, which is significantly faster than other state-of-the-art competitors including RANSAC, GNC and ADAPT.

\section{Related Work}

\subsection{Naive Estimation Methods without Outliers}\label{sec-minimal}

\textbf{Minimal Solvers.} Minimal solvers can make estimates with the fewest measurements possible, and its typical examples include: Horn's triad-based method~\cite{horn1987closed} for point cloud registration, and the 5-point solver~\cite{nister2004efficient} for two-view geometry. However, due to the lack of redundant measurements to refine the solution, minimal solvers may be easily affected by noise (e.g. as discussed in~\cite{chum2003locally}). Moreover, some problems are unsuitable to be solved by minimal solvers because they have too many variables to solve (e.g. category-level perception that involves solving multiple shape parameters) and this will make the dimension (number of measurements required) of the minimal solvers too high and hence tremendously slowing down the convergence when running with RANSAC.

\textbf{Non-minimal Solvers.} Non-minimal solvers takes an arbitrary number of measurements as inputs (on the condition that the problem is overdetermined) to make the optimal estimation by solving least-squares optimization under the assumption of Gaussian noise. Typical examples include: Arun's SVD method~\cite{arun1987least} for point cloud registration, Semi-Definite relaxations for 3D registration~\cite{briales2017convex} and category-level perception~\cite{shi2021optimal}, as well as Sum-of-Squares relaxations for 2D-3D shape alignment \& reconstruction~\cite{yang2020graduated,yang2020perfect}. Though non-minimal solvers can minimize the effect of noise and yield optimal estimates, they usually require relatively long runtime compared to minimal solvers (e.g. due to the relaxation methods on high-degree polynomial optimization problems). Thus, once combined with the outlier rejection frameworks (e.g. GNC, ADAPT) where plenty of iterations are needed, the time-efficiency issue appears critical.

\subsection{Outlier-Robust Estimation Methods}

\textbf{Consensus Maximization.} It seeks to find the model that corresponds to the largest number of measurements with residual errors below the inlier threshold in order to achieve robust estimation. RANSAC~\cite{fischler1981random} is the most popular consensus maximization method that randomly samples small measurement subsets to make minimal model estimates and finds the best model enabling the largest measurement consensus set. Additional techniques such as local optimization~\cite{chum2003locally,lebeda2012fixing} or measurement ranking~\cite{chum2005matching} have been applied to improve RANSAC. But RANSAC methods' natural exponential runtime with the growth of outlier ratio and problem dimension confines their usage in many practical situations. Branch-and-Bound~\cite{li2009consensus} is another maximum consensus framework that returns optimal results by searching in the parameter space, but it also suffers from the worst-case exponential time w.r.t. problem size. Also, ADAPT~\cite{tzoumas2019outlier} can reject outliers by alternating between non-minimal estimation and measurement trimming with a gradually decreasing residual threshold. But it often requires dozens to hundreds of iterations to converge, and considering the slow runtime of some non-minimal solvers, time-efficiency is still a notable issue.

\textbf{M-Estimation.} M-estimation employs robust cost functions to diminish the weights of outliers. Earlier local M-estimation methods~\cite{mactavish2015all} usually require an initial guess and then conduct iterative optimization for convergence, but they are prone to get stuck in local minimum if the initialization is bad. Graduated Non-Convexity (GNC) is presented to avoid the need of the initial guess as a global estimator, which is first used in~\cite{zhou2016fast} for robust point cloud registration. Then, GNC is promoted as a general-purpose estimator for diverse perception problems~\cite{yang2020graduated} and its accelerated version GNC-IRLS~\cite{peng2023convergence} is also supplemented more currently. But GNC typically needs 30--50 iterations to converge while GNC-IRLS also requires 7-20+ iterations in our experiments, thus confronted with the same runtime issue as ADAPT.

\begin{figure*}[ht]
\centering

\setlength\tabcolsep{0pt}
\addtolength{\tabcolsep}{0pt}

\begin{tabular}{cccccc}
\footnotesize{Iteration 1} & \footnotesize{Iteration 2} & \footnotesize{Iteration 3} & \footnotesize{Iteration 4}
& \footnotesize{Iteration 5} & \footnotesize{Iteration 6 (Converged)} \\
\begin{minipage}[t]{0.16\linewidth}
\centering
\includegraphics[width=1\linewidth]{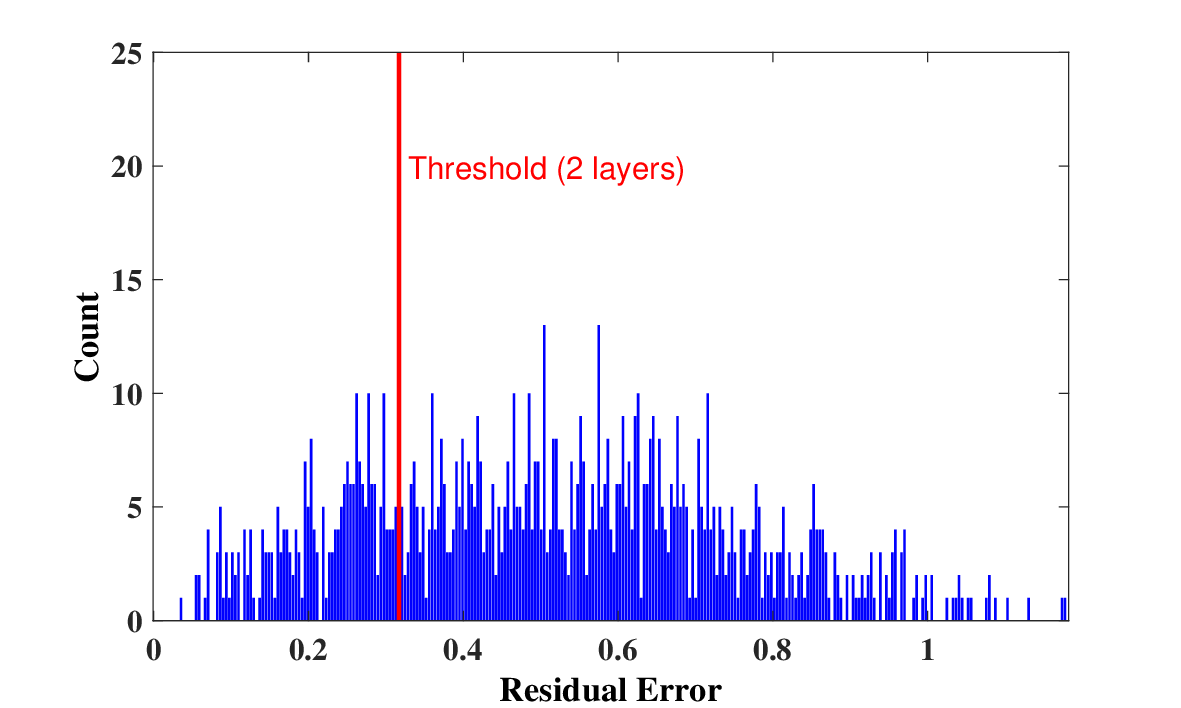}
\end{minipage}
&
\begin{minipage}[t]{0.16\linewidth}
\centering
\includegraphics[width=1\linewidth]{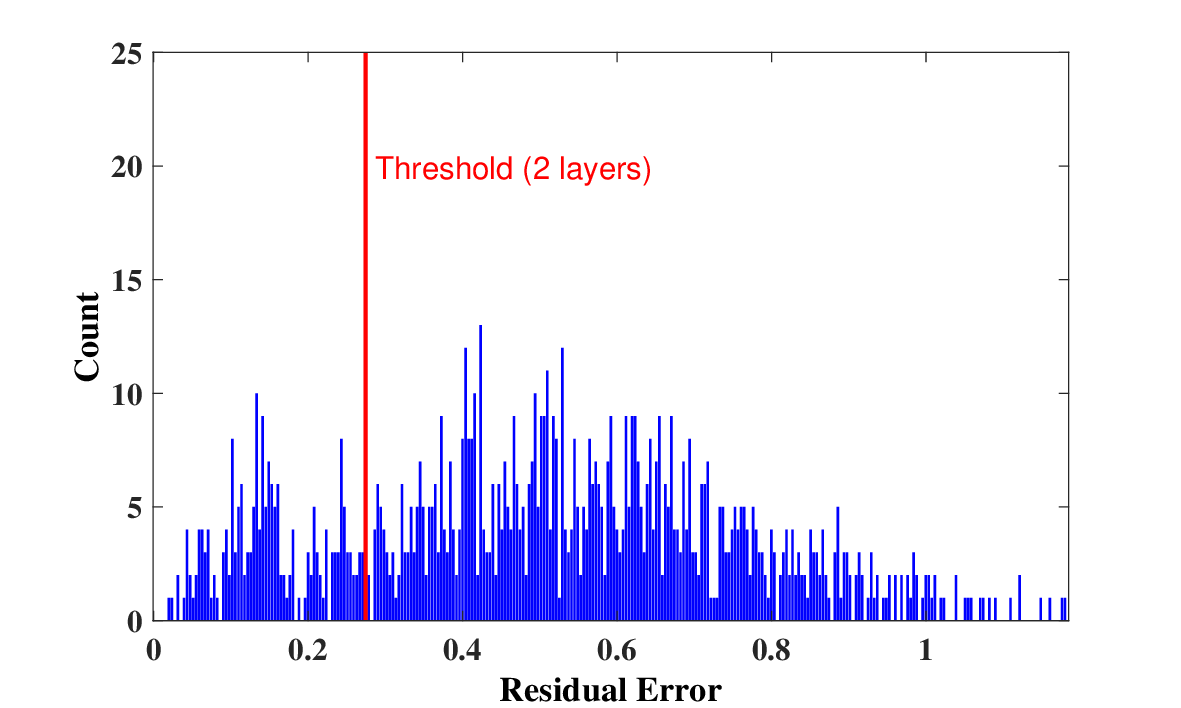}
\end{minipage}
&
\begin{minipage}[t]{0.16\linewidth}
\centering
\includegraphics[width=1\linewidth]{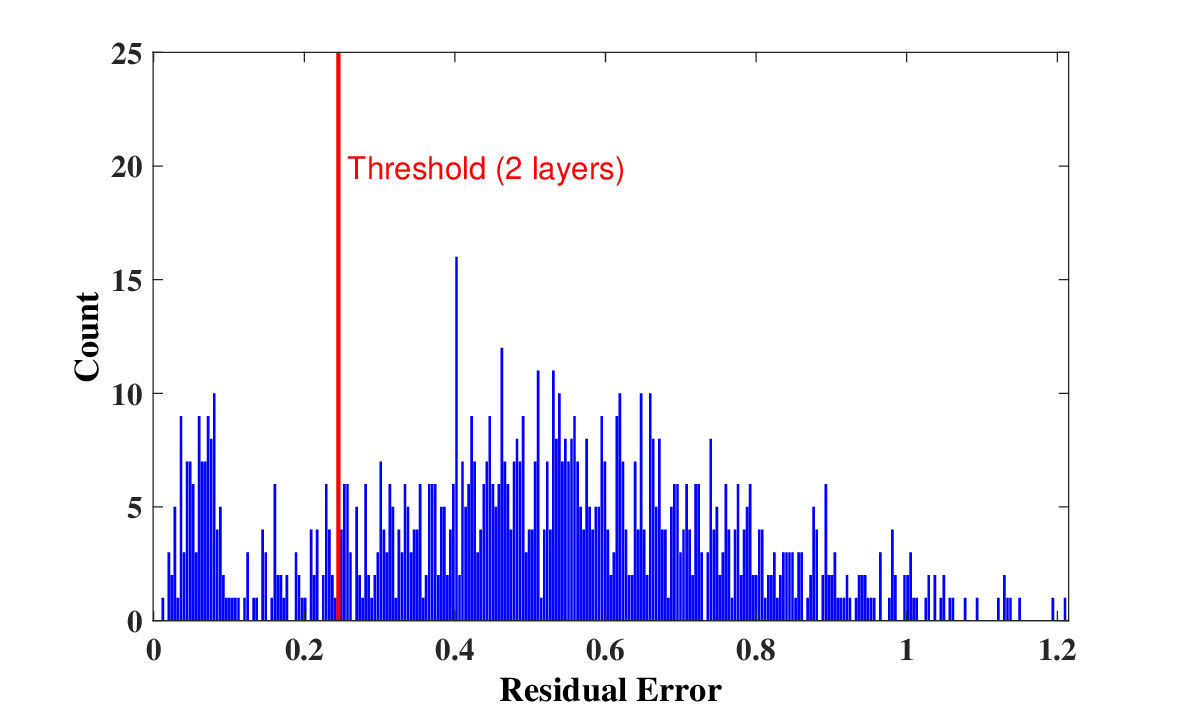}
\end{minipage}
&
\begin{minipage}[t]{0.16\linewidth}
\centering
\includegraphics[width=1\linewidth]{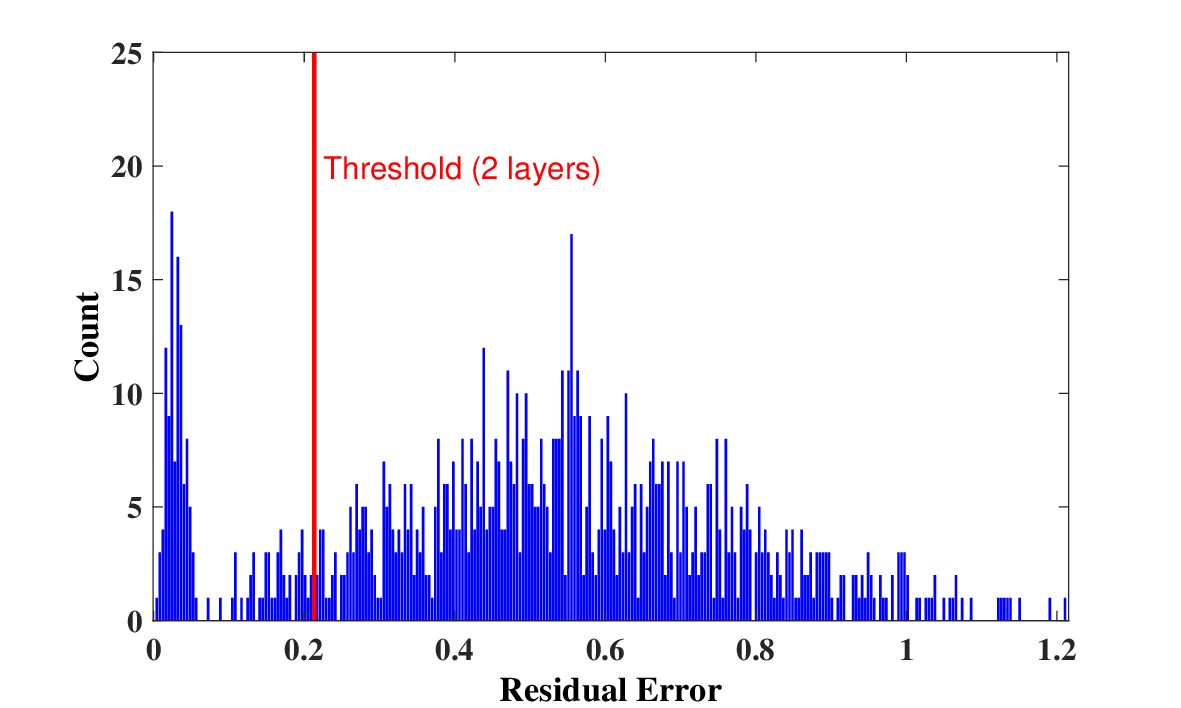}
\end{minipage}
&
\begin{minipage}[t]{0.16\linewidth}
\centering
\includegraphics[width=1\linewidth]{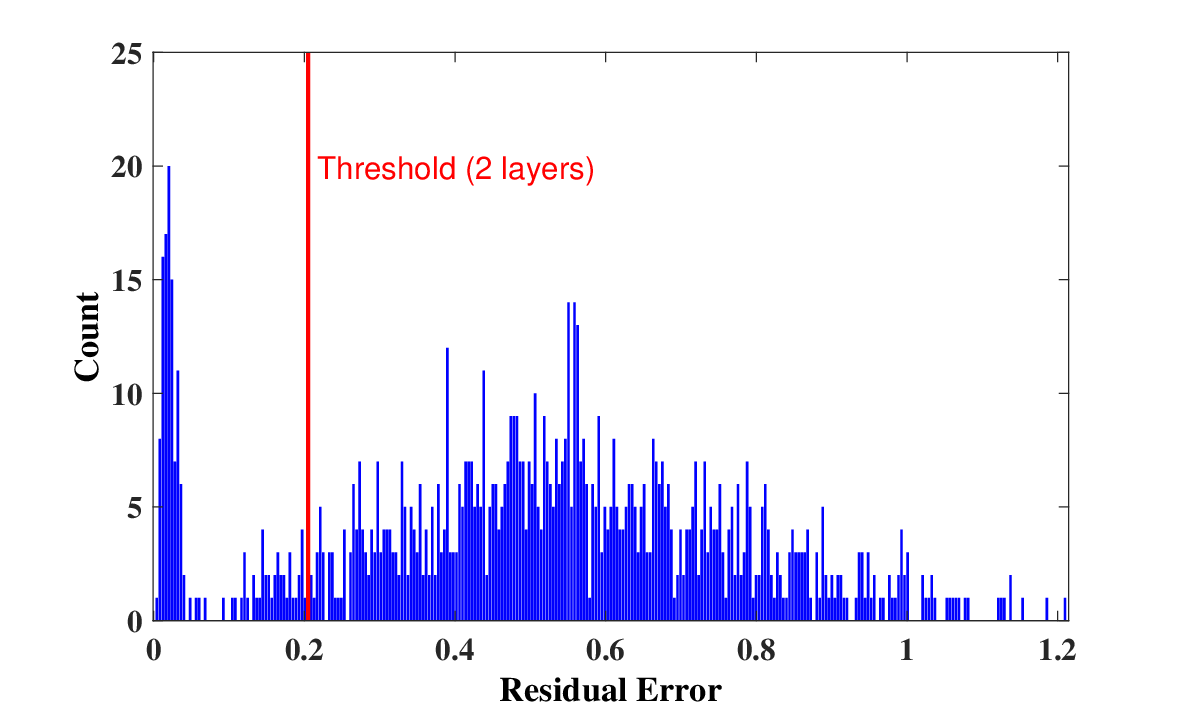}
\end{minipage}
&
\begin{minipage}[t]{0.16\linewidth}
\centering
\includegraphics[width=1\linewidth]{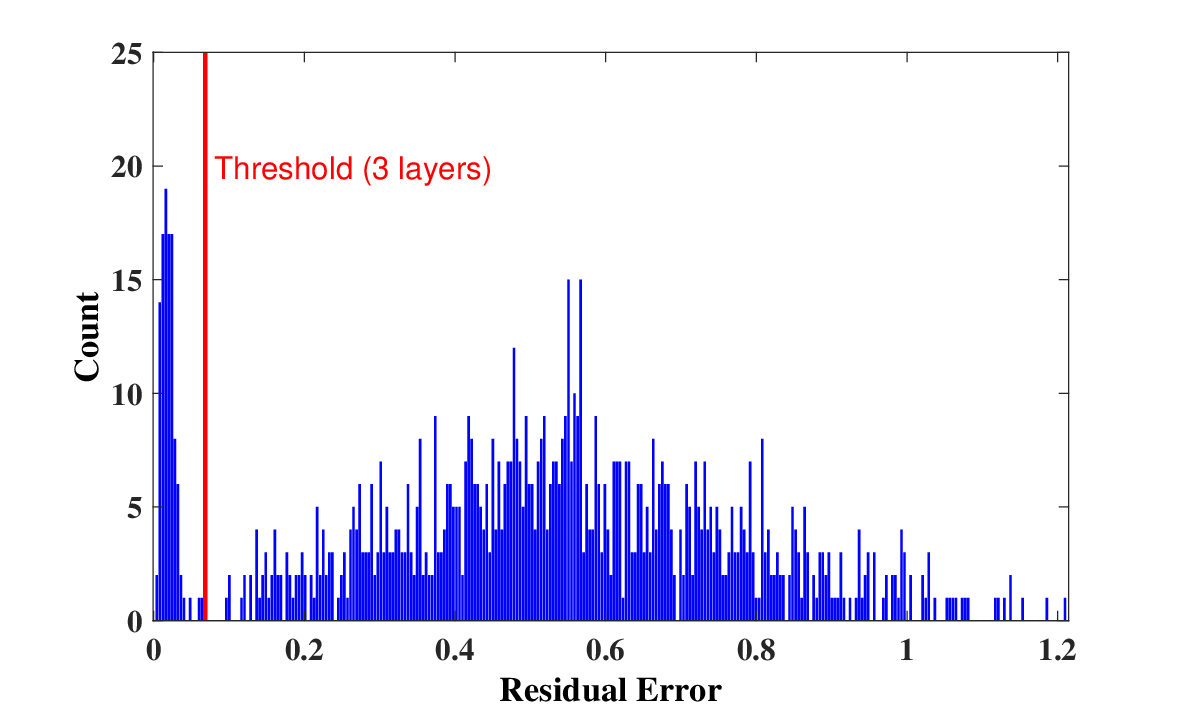}
\end{minipage}
\end{tabular}

\vspace{-1mm}
\caption{Illustration of the residual-error histograms with labeled thresholds in our TIVM algorithm at different iterations.}
\label{hist-demo}
\vspace{-5mm}
\end{figure*}

\section{Preliminaries for Geometric Perception}

We first provide a general definition for geometric spatial perception problems as in~\cite{yang2020graduated}. For a geometric estimation problem where its model (a.k.a. variables to solve) is represented as $\boldsymbol{x} \in \mathcal{X}$ (here $\mathcal{X}$ is the feasible domain), its measurements are denoted as $\boldsymbol{m}_{i},\,i\in\mathcal{N}=\{1,2,\dots,N\}$, and the residual function that measures the difference (non-negative error) between the measurements computed with current model $\boldsymbol{x}$ and the actual measurements $\boldsymbol{m}_{i}$ is formulated as $Re\left(\boldsymbol{m}_{i}, \boldsymbol{x}\right)$ (also abbreviated as $Re_i$). When there exits no outlier in the measurements, this problem can be optimally solved by the following minimization formulation: 
\begin{equation}\label{PF}
\min _{\boldsymbol{x} \in \mathcal{X}} \sum_{i=1}^{N} Re^{2}\left(\boldsymbol{m}_{i}, \boldsymbol{x}\right).
\end{equation}

For example, if the estimation problem is point cloud registration, $\boldsymbol{x}$ should denote the rigid transformation, $\mathcal{X}$ should be $SE(3)$, and $\boldsymbol{m}_{i}$ would be the putative correspondences matched between point clouds. 

However, in reality, measurements $\boldsymbol{m}_{i}$ are often corrupted by outliers, so we need a outlier-robust formulation for this estimation problem which can be represented as a consensus maximization problem (\cite{tzoumas2019outlier,sun2022trivoc}) such that
\begin{equation}\label{CM}
\begin{gathered}
\underset{\mathcal{I}\subset \mathcal{N}}{\max}\, |\mathcal{I}|, \\
s.t. \,|Re\left(\boldsymbol{m}_{i}, \boldsymbol{x}^{\star}\right)| \leq \gamma\,\,(\forall i\in\mathcal{I})\, ,
\end{gathered}
\end{equation}
where the goal is to find the optimal model $\boldsymbol{x}^{\star}$ that enables as many measurements $\boldsymbol{m}_{i}$ as possible to satisfy the condition that their residual errors $Re_i$ are lower than the user-defined inlier threshold $\gamma$ (also known as the noise bound) that serves as the criterion to differentiate inliers from outliers. 

For robust perception problems without inlier-noise statistics, threshold $\tau$ would be unknown, which greatly increases the difficulty of robust estimation; but fortunately, we manage to design a novel paradigm that uses maximum intra-class variance to achieve noise-statistics-free robust estimation, as rendered in the following section.

\section{Methodology}

\textbf{Motivation.} As discussed in~\cite{tzoumas2019outlier}, for the optimal model $\boldsymbol{x}^{\star}$, its measurement residuals $Re_i$ should naturally form 2 relatively distinct groups, one with low magnitude (as inliers) and the other with high magnitude (as outliers). Thus, our idea is to design an iterative dynamic thresholding framework to gradually approximate the optimal model by achieving 2 groups of measurements with the maximum intra-group variance on their residual errors in each iteration. And thanks to this framework, one huge advantage of the proposed algorithm is that: even when the inlier-noise statistics are not given, it could still return robust (reasonably accurate) estimation results. 

We provide the pseudocode: Algorithm~\ref{algo-TIVM*} for the noise-statistics-free scenario (without known inlier threshold $\tau$) in advance, for the convenience of the subsequent elaboration.

\subsection{Intra-Class Variance Maximization for Dynamic Residual Thresholding}\label{sec-IVM}


Intra-class variance maximization is an efficacious approach to classify a set of data into multiple groups in a way that the largest possible variance between classes is achieved, and its most popular application is in Otsu's image thresholding method~\cite{otsu1979threshold}. In this work, we employ this technique to accomplish dynamic thresholding on the residual errors, trying to distinguish inliers from outliers. 


A prerequisite for this technique is to create a series of consecutive \textit{intervals} to encompass all the residual errors so as to construct a `residual histogram', as exemplified in Fig.~\ref{hist-demo}. Given that all residual errors are positive, we set the maximum residual error as the upper-bound of all intervals such that $D^{max}=max(Re_i), i\in\mathcal{N}$, and then divide $D^{max}$ into $L$ consecutive small intervals where each interval has the same length of $\Delta D=\frac{D^{max}}{L}$ (typically, we set $L=300$ in practice). Now there exist $L$ intervals given by: $\left(0,\Delta D\right]$, $\left(\Delta D, 2\Delta D\right]$, $\left(2\Delta D, 3\Delta D\right]$, $\dots$, $\left(D^{max}-\Delta D,D^{max}\right]$. 

Then, we let $n_l\, (l=1,2,\dots,L)$ denote the number of residual errors that fall into the $l_{th}$ interval, so apparently $\sum_{l=1}^{L} n_l=N$. Based on the procedures derived in~\cite{otsu1979threshold}, we normalize $n_l$ as $p_l=\frac{n_l}{N}$, so the probability for a residual error to be lower than $k\cdot\Delta D$ can be represented as:
\begin{equation}\label{Pk-compute}
P_k=\sum_{l=1}^k\, p_l,
\end{equation}
where $k$ is an integer ranging from $1$ to $L$. The cumulative mean of residual errors lower than $k\cdot\Delta H$ can be given by:
\begin{equation}
\mu_k=\sum_{l=1}^k\, l \cdot p_l. 
\end{equation}
Letting $k=L$, we can have the mean of all residual errors:
\begin{equation}\label{uG-compute}
\bar{\mu}=\mu_{L}=\sum_{l=1}^{L}\, l \cdot p_l.
\end{equation}
Up to now, the intra-class variance for a certain $k\,(\text{where }k=1,2,\dots,L)$ can be computed as:
\begin{equation}\label{variance-compute}
\sigma_k=\frac{\left(\bar{\mu} P_{k}-\mu_{k}\right)^{2}}{P_{k}\left(1-P_{k}\right)}.
\end{equation}
As a result, the best threshold $T^\star$ can be easily solved by:
\begin{equation}\label{T-compute}
T^\star=\Delta D\cdot k^\star=\Delta D \cdot\underset{k=1,2,\dots,L}{\arg\max}(\sigma_k).
\end{equation}

With the best $T^\star$, we can dynamically separate all the measurements into 2 groups based on their residual-error histogram as: a lower-magnitude group $\mathcal{G}^\alpha$ with residual errors no larger than $T^\star$ ($\forall i\in\mathcal{G}^\alpha, Re_i\leq T^\star$), and a higher-magnitude group $\mathcal{G}^\beta$ with residual errors larger than $T^\star$ ($\forall i\in\mathcal{G}^\beta, Re_i>T^\star$). 


However, when outliers are prevalent among measurements (e.g. $>50\%$), grouping the measurements based on the routine above may still include outliers into the lower-magnitude group $\mathcal{G}^\alpha$, hindering us from correctly sifting out the true inliers. Therefore, we need to further perform multiple layers of dynamic thresholding to enhance the inlier confidence of $\mathcal{G}^\alpha$. 

\subsection{Multi-Layered Thresholding for Outlier Rejection}\label{sec-thresholding}

Once we complete the first thresholding over all measurements with threshold $T^\star_1$ to obtain group $\mathcal{G}_1^\alpha$ (where subscript $j$ denotes that $\mathcal{G}_j^\alpha$ and $T^\star_j$ are obtained in the $j_{st}$ layer of thresholding), we can conduct another layer of dynamic thresholding on $\mathcal{G}_1^\alpha$, that is, treating all measurements in the current lower-magnitude group $\mathcal{G}_1^\alpha$ as the input and applying~\eqref{Pk-compute}--\eqref{T-compute} again to further divide $\mathcal{G}_1^\alpha$ into 2 groups: $\mathcal{G}_2^\alpha$ and $\mathcal{G}_2^\beta$ with the new threshold $T^\star_2$ estimated from~\eqref{T-compute}. Thus, assuming the residual errors are computed over the optimal model $\boldsymbol{x}^\star$, if we conduct $m$ layers of such 2-group thresholding where $m$ is adequate, the final low-magnitude group in the last layer, denoted as $\mathcal{G}_m^\alpha$, should be a pure-inlier (outlier-free) group. \,(This multi-layered thresholding process corresponds to \textit{Lines 5--12} in Algorithm~\ref{algo-TIVM*}.) 

But unfortunately, the model we use for estimation could not be the optimal model since $\mathcal{G}_m^\alpha$ is not supposed to contain pure inliers. Thus, in this circumstance, we introduce an iterative optimizing framework where in each iteration, the measurements in $\mathcal{G}_m^\alpha$ are repeatedly updated and adopted for non-minimal estimation and residual computing in the next iteration. In this framework, model $\boldsymbol{x}$ and group $\mathcal{G}_m^\alpha$ are gradually optimized to approach the optimal (pure-inlier) solutions via iterations of the multi-layered dynamic thresholding procedure as described above.\, (This process corresponds to \textit{Lines 2--4 \& 21} in Algorithm~\ref{algo-TIVM*}.) 


\begin{algorithm}[t]
\caption{{TIVM$^\circ$}\, (Noise-statistics-free Version)}
\label{algo-TIVM*}
\SetKwInOut{Input}{\textbf{Input}}
\Input{measurements $\{\boldsymbol{m}_{i}\}_{i\in\mathcal{N}}$\;}
Set $\mathcal{C}^0\leftarrow\mathcal{N}$, $L\leftarrow300$, $m\leftarrow2$， $checkConv\leftarrow false$, and $t_{max} \leftarrow 100$\;
\For{$t=1:t_{max}$}{
Solve model $\boldsymbol{x}^t$ using the non-minimal solver with measurements in set $\mathcal{C}^{t-1}$\;
Compute the residual errors $Re_i$ w.r.t. all the measurements $\boldsymbol{m}_i$ ($i\in\mathcal{N}$) using $\boldsymbol{x}^t$\;
$D^{max}\leftarrow\max(Re_i)$ and $\Delta D\leftarrow \frac{D^{max}}{L}$\;
Set $L$ intervals and count $n_l$ ($l\in\{1,2,\ldots,L\}$)\;
$\mathcal{G}^\alpha_0\leftarrow \mathcal{N}$ and $T^\star_0\leftarrow L$\;
\For{j=1:m}{
$\forall l\in\{1,2,\ldots,\frac{T^\star_{j-1}}{\Delta D}\}$, compute $p_l\leftarrow \frac{n_l}{|\mathcal{G}_{j-1}^\alpha|}$\;
$\forall k\in\{1,2,\ldots,\frac{T^\star_{j-1}}{\Delta D}\}$, update $P_k$, $\mu_k$, $\bar{\mu}$ and $\sigma_k$ according to~\eqref{Pk-compute}--\eqref{variance-compute}\; 
Solve threshold $T^\star_{j}$ based on~\eqref{T-compute} to separate $\mathcal{G}_{j-1}^\alpha$ into 2 groups: $\mathcal{G}_{j}^\alpha$ and $\mathcal{G}_{j}^\beta$\;
}
\If{$checkConv$ \textup{and} $\frac{\left|\bar{Re}-\frac{1}{N}\sum_{i=1}^N Re_i\right|}{\bar{Re}}\leq 10^{-3}$} {
\textbf{break}
}
$checkConv\leftarrow false$, $T^{t}\leftarrow T^\star_{m}$ and $\mathcal{C}^{t}\leftarrow\mathcal{G}_{m}^\alpha$\;
\If{$|T^t-T^{t-1}|\leq\Delta D$}{
$m\leftarrow m+1$ and $checkConv\leftarrow true$\;
$\bar{Re}=\frac{1}{N}\sum_{i=1}^N Re_i$\;
}
}
\Return optimal model $\boldsymbol{x}^\star\leftarrow \boldsymbol{x}^t$ and inlier set $\mathcal{C}^{t}$\;
\end{algorithm}

\subsection{Self-Adaptive Layer Number Tuning}\label{sec-adaptive}

Nonetheless, there exists a challenge that: it is hard for the users to preset the appropriate layer number $m$ for dynamic thresholding. If $m$ is too small, outliers may be included in $\mathcal{G}_m^\alpha$; if $m$ is too large, the measurement samples for non-minimal estimation would be too sparse, which may compromise the estimation accuracy. 

To resolve this issue, we design a self-adaptive mechanism for automatically tuning the layer number $m$. Specifically, we first start dynamic thresholding with a small value of $m$ (e.g. $m=2$ in practice), and when the iterative optimizing framework converges, we would examine if the current group $\mathcal{G}_m^\alpha$ is good enough. This could be done by checking whether the model estimated by the converged $\mathcal{G}_m^\alpha$ set is similar enough to the model estimated by the $\mathcal{G}_{m+1}^\alpha$ after one more iteration with an incremented layer number: $m+1$. If these 2 models have close enough average residual errors (which indicates the models are similar enough), then the current converged $\mathcal{G}_m^\alpha$ should be a pure-inlier group. But if not, meaning that the current layer number is insufficient for separating the pure inliers, then the iteration should continue with a new incremented layer number: $m\leftarrow m+1$. \,(This mechanism corresponds to \textit{Lines 13--20} in Algorithm~\ref{algo-TIVM*}). 

So far, Algorithm~\ref{algo-TIVM*} provides the full algorithm (that we name TIVM$^\circ$) for robust estimation in the noise-statistics-free scenario. An example illustration of the residual-error histograms at all the iteration of the proposed algorithm is shown in Fig.~\ref{hist-demo}, where we can observe that: as the iteration goes on, the distribution of the residual errors changes from 1 cluster with hardly any clear separation boundary into 2 distinct groups including: an inlier group with low residual magnitude and low variance (since inliers are consistent) plus an outlier group with high residual magnitude and high variance (since outliers are random).

\subsection{TIVM with Noise-Statistics Information}

In some cases, the inlier-noise statistics information is known by the user, so we could further boost the estimation accuracy with the aid of the given inlier threshold $\tau$, and the new version is shown in Algorithm~\ref{algo-TIVM} (that we name TIVM). The main structure of the algorithm remains unchanged. The difference is that: we could add another convergence condition that: when the current threshold $T^\star_m$ is smaller than $2\cdot\tau$ (indicating that the current model is good enough to differentiate inliers from outliers in 2 sufficiently distinct clusters), we should stop the iterations and then estimate the optimal model $\boldsymbol{x}^\star$ directly using the measurements with residual errors lower than $\tau$.

\begin{algorithm}[t]
\caption{{TIVM}\, (Noise-statistics Version)}
\label{algo-TIVM}
\SetKwInOut{Input}{\textbf{Input}}
\Input{measurements $\{\boldsymbol{m}_{i}\}_{i\in\mathcal{N}}$, inlier threshold $\tau$\;}
$\ldots\ldots\ldots$\,\, (\textit{Lines 1--15} in Algorithm~\ref{algo-TIVM*}) \\
\If {$T^\star_m\leq 2\cdot\tau$}{
\textbf{break}
}
$\ldots\ldots\ldots$\,\, (\textit{Lines 16--21} in Algorithm~\ref{algo-TIVM*}) \\
Find all the measurements whose residual errors satisfy: $Re_i\leq\tau$ and add them to the inlier set $\mathcal{C}^\star$\;
Solve optimal model $\boldsymbol{x}^\star$ with measurements in $\mathcal{C}^\star$\;
\Return optimal model $\boldsymbol{x}^\star$ and full inlier set $\mathcal{C}^\star$\;
\end{algorithm}

Note that Algorithm~\ref{algo-TIVM} only serves to lower the estimation errors by finding the full inlier set using the ground-truth inlier-noise threshold $\tau$, but theoretically it cannot change the outlier-robustness because it is determined by whether the final low-magnitude group $\mathcal{G}_{m}^\alpha$ contains outliers (which completely relies on Algorithm~\ref{algo-TIVM*}).

\section{Experiments}

We evaluate our estimator TIVM (and the noise-statistics-free version TIVM$^\circ$) in 3 geometric perception problems: rotation averaging, point cloud registration and category-level perception. In each problem, we benchmark our estimator against state-of-the-art general-purpose robust solvers including: (i) GNC-TLS and GNC-GM~\cite{yang2020graduated} with control parameter $\mu=1.4$ (according to~\cite{yang2020graduated}) and 100 maximum iterations, (ii) GNC-IRLS~\cite{peng2023convergence} with all parameters set as in source code of~\cite{yang2020graduated} with 100 maximum iterations, (iii) ADAPT~\cite{tzoumas2019outlier} where the discount ratio is set to 0.99 as suggested in~\cite{tzoumas2019outlier} with 200 maximum iterations, and (iv) other problem-specific specialized solvers. In addition, we test with unknown inlier-noise statistics where inlier threshold $\tau$ is unavailable and TIVM$^\circ$ is compared against the state-of-the-art minimally-tuned solvers: GNC-MinT and ADAPT-MinT in~\cite{antonante2021outlier} with 1000 maximum iterations. Experiments are conducted in Matlab on a laptop having an i9-12900H CPU and 32GB RAM with single thread.

\begin{figure}[t]
\centering

\setlength\tabcolsep{2pt}
\addtolength{\tabcolsep}{0pt}

\begin{tabular}{cccc}

\begin{minipage}[t]{0.5\linewidth}
\centering
\includegraphics[width=1\linewidth]{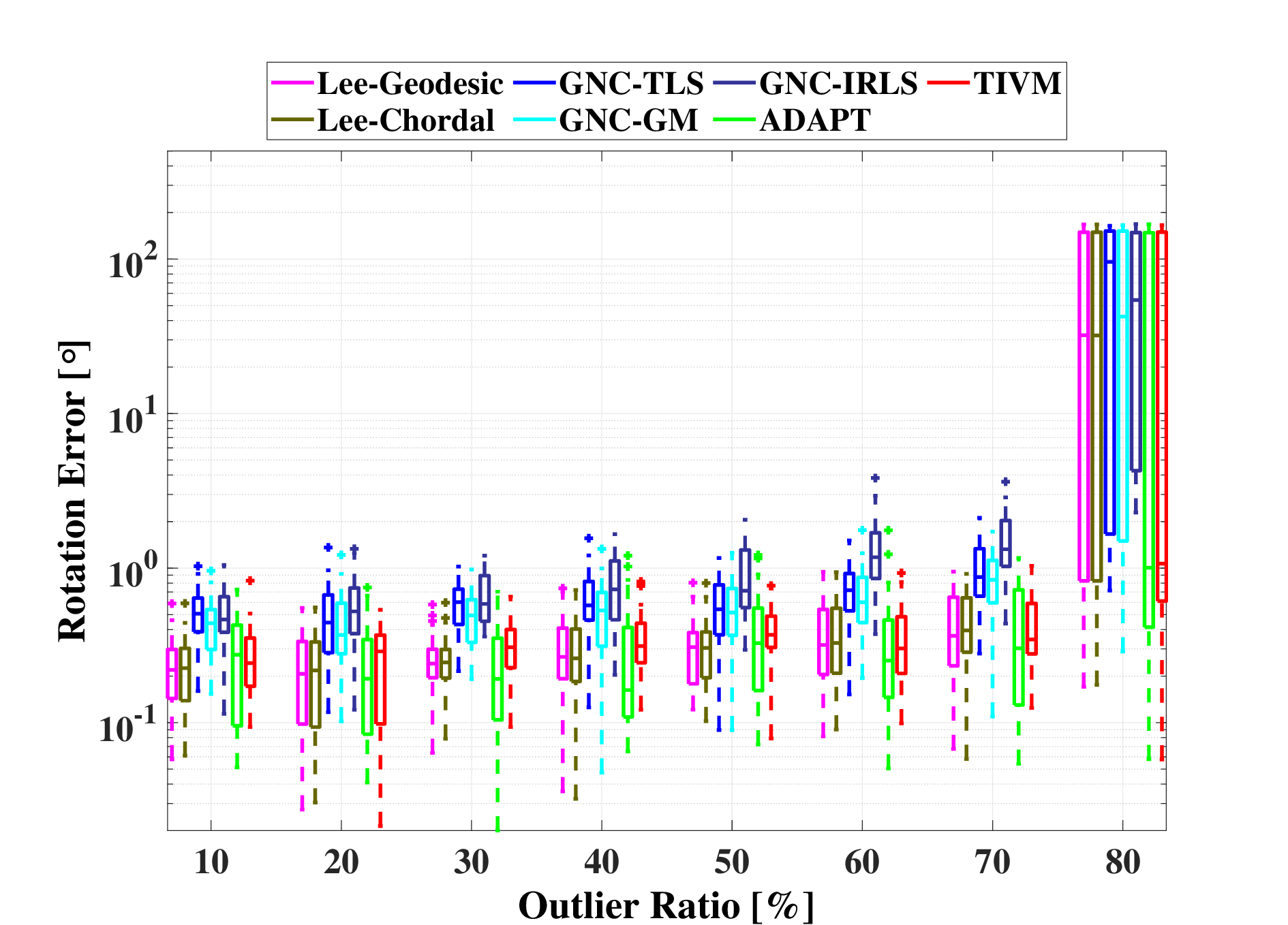}
\end{minipage}
&
\begin{minipage}[t]{0.5\linewidth}
\centering
\includegraphics[width=1\linewidth]{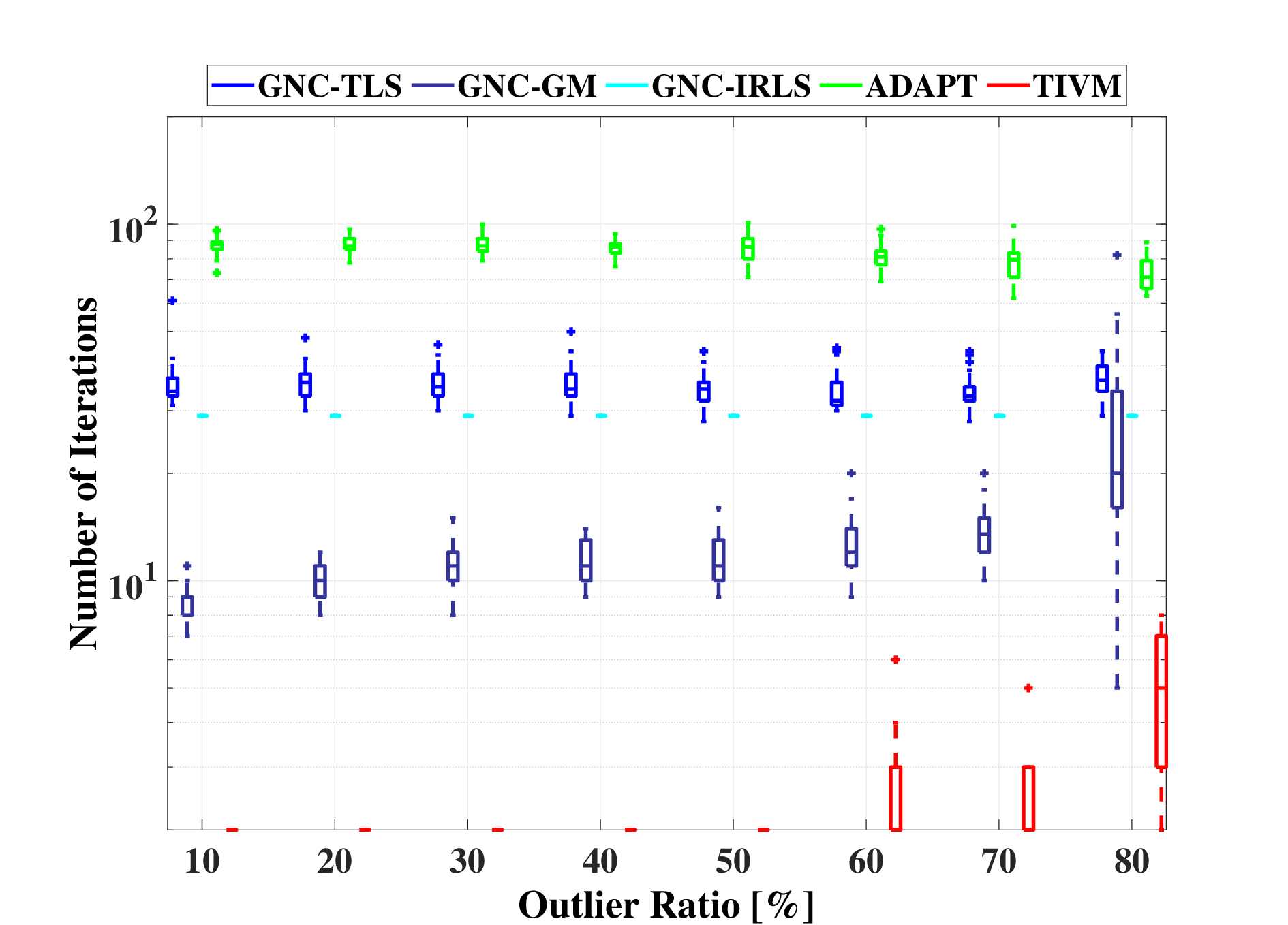}
\end{minipage}

\end{tabular}

\vspace{-1mm}
\caption{Benchmarking on robust rotation averaging.}
\label{Benchmarking-RA}
\vspace{-1mm}
\end{figure}

\begin{figure}[t]
\centering

\setlength\tabcolsep{1pt}
\addtolength{\tabcolsep}{0pt}

\begin{tabular}{cc}

\begin{minipage}[t]{0.5\linewidth}
\centering
\includegraphics[width=1\linewidth]{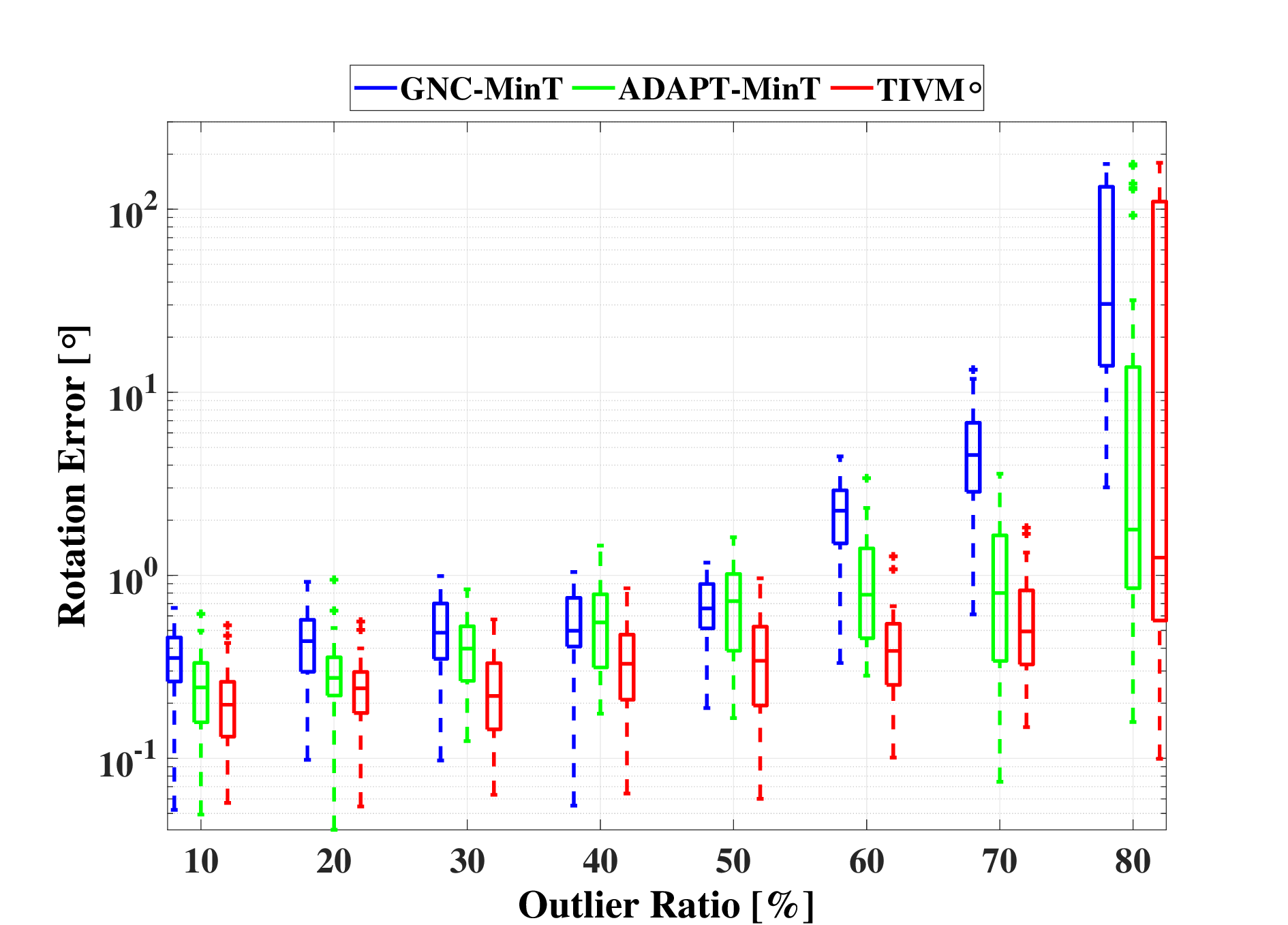}
\end{minipage}
&
\begin{minipage}[t]{0.5\linewidth}
\centering
\includegraphics[width=1\linewidth]{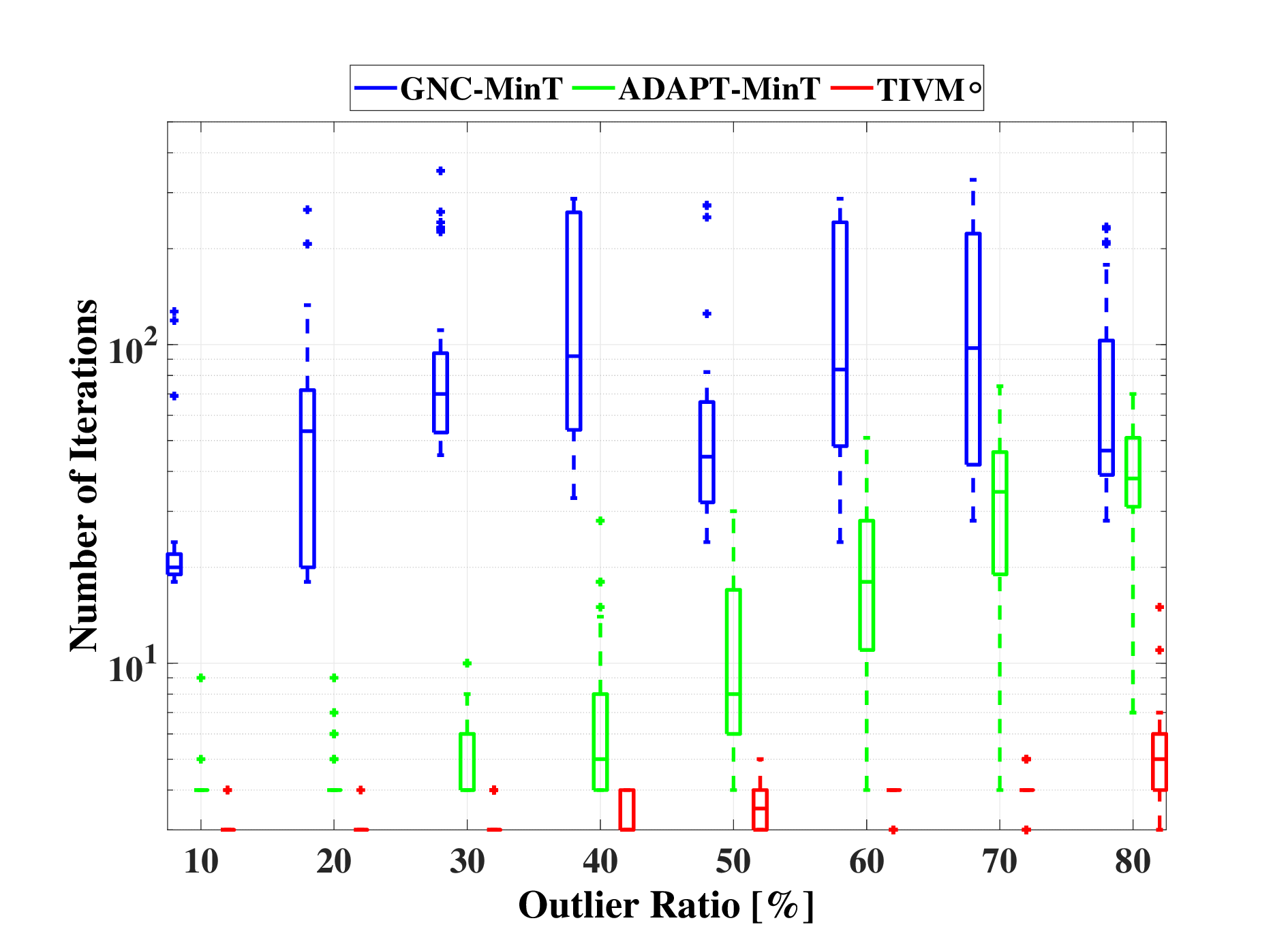}
\end{minipage}

\end{tabular}

\vspace{-1mm}
\caption{Benchmarking on robust rotation averaging without inlier-noise statistics.} 
\label{Benchmarking-RA-NS-free}
\vspace{-1mm}
\end{figure}

\subsection{Rotation Averaging}

Robust rotation averaging aims to find the best rotation $\boldsymbol{R}^{\star}\in SO(3)$ from a series of noisy rotation estimates w.r.t. the ground-truth value: $\boldsymbol{R}_i\in SO(3),\,i=1,2,\dots,N$, which are potentially corrupted by outliers. The residual error can be defined as: $Re_i=G(\boldsymbol{R}_i,\boldsymbol{R}^\star)$, where `$G(\cdot\,,\,\cdot)$' represents the geodesic distance~\cite{hartley2013rotation} between rotations.

\begin{figure}[b]
\centering

\setlength\tabcolsep{1pt}
\addtolength{\tabcolsep}{1pt}

\begin{tabular}{cc}

\multicolumn{1}{c}{\footnotesize{(a) 60\% Outliers}} & 
\multicolumn{1}{c}{\footnotesize{(b) 90\% Outliers}}\\
\begin{minipage}[t]{0.42\linewidth}
\centering
\includegraphics[width=1\linewidth]{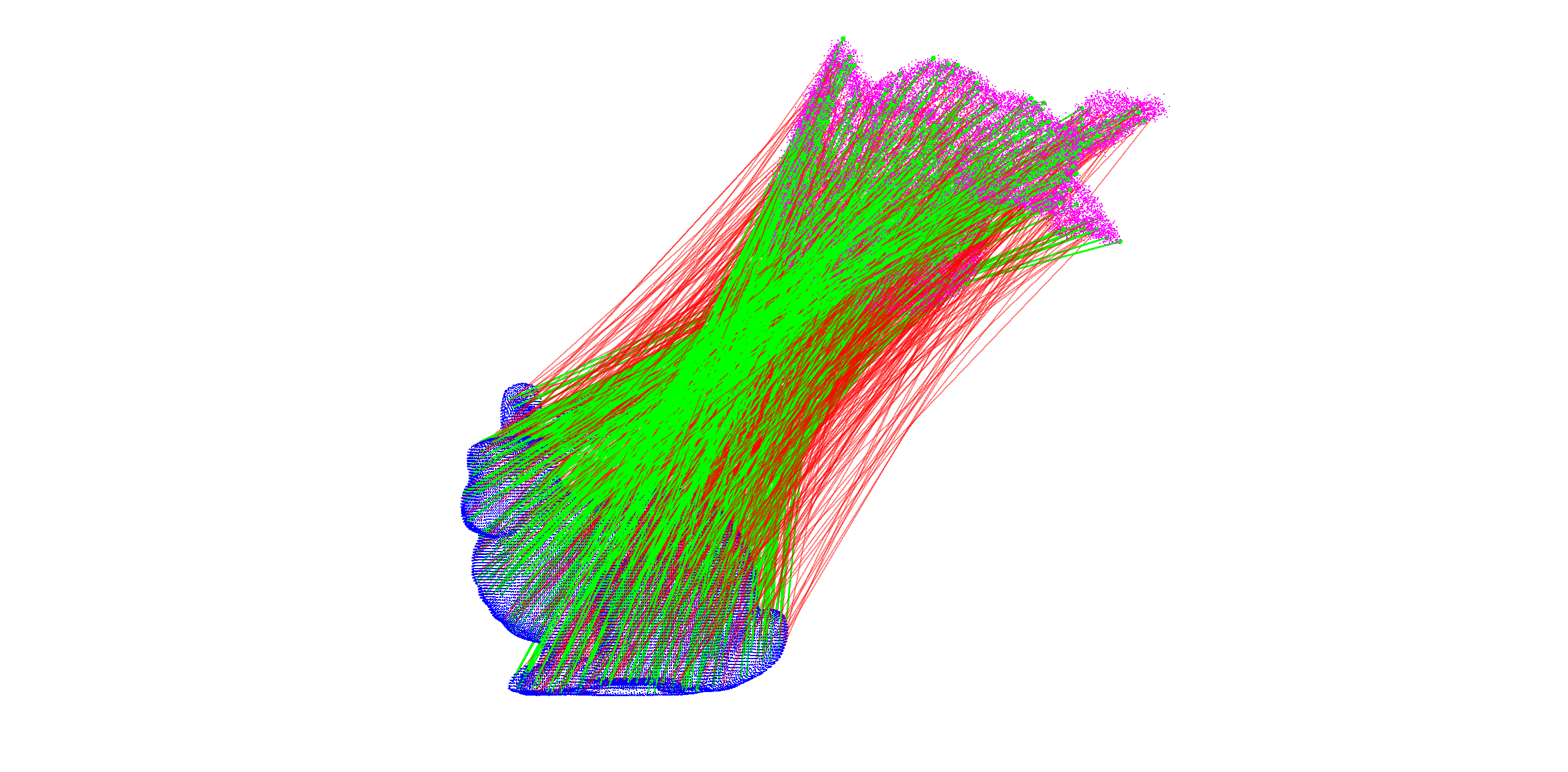}
\end{minipage}
&
\begin{minipage}[t]{0.42\linewidth}
\centering
\includegraphics[width=1\linewidth]{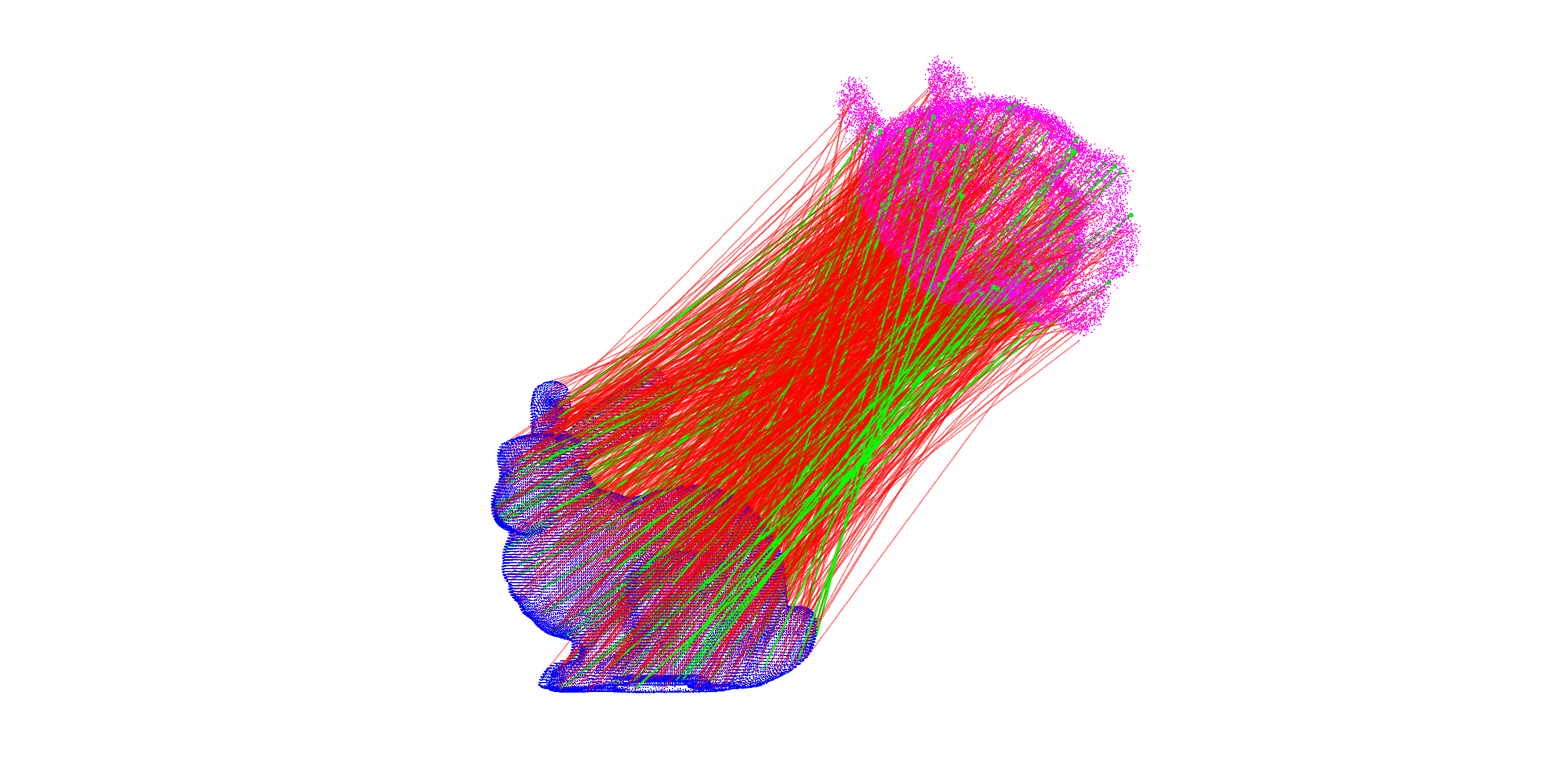}
\end{minipage}
\end{tabular}

\vspace{-1mm}
\caption{Demonstration of the experimental setup of robust point cloud registration. Green and red lines denote inliers and outliers, respectively.}
\label{demo-PCR}
\vspace{-1mm}
\end{figure}

\textbf{Setup.} In each run, we obtain a random rotation $\boldsymbol{R}_{gt}\in SO(3)$ as the ground truth and then generate $N=100$ noisy rotations $\boldsymbol{R}_i$ as the input measurements such that $\boldsymbol{R}_i=\boldsymbol{R}_{gt}Exp(\eta_i\boldsymbol{e}_i)$ where $\eta_i$ is a random isotropic Gaussian noise angle with standard deviation of $\sigma=5^\circ$, $\boldsymbol{e}_i$ is a random unit-norm 3D vector and $Exp(\cdot)$ is the exponential map. To create outliers, we replace 10--80\% of the rotations in $\{\boldsymbol{R}_i\}_{i=1}^N$ with random rotations. For specialized solvers in benchmarking, we adopt the state-of-the-art Chordal-median and Geodesic-median methods~\cite{lee2020robust}. The inlier threshold is set as $\tau=3\sigma$. In GNC, we use the weighted L1-chordal mean for weighted non-minimal rotation estimation; in ADAPT and TIVM, we use the L1-chordal median~\cite{lee2020robust} as the non-minimal solver. The benchmarking results are based on 30 Monte Carlo runs, as shown in Fig.~\ref{Benchmarking-RA}.


From Fig.~\ref{Benchmarking-RA}, we see that: (1) all the solvers tolerate 70\% outliers but fail at 80\% outliers, (2) among general-purpose solvers, our TIVM is the most efficient solver that requires fewer than 10 iterations, GNC-IRLS is the second fastest solver mostly with 7--20 iterations, and GNC-TLS, GNC-GM and ADAPT need 30--100 iterations, and (3) TIVM has higher estimation accuracy than the 3 GNC solvers.

\textbf{Unknown Noise-Statistics Tests.} We perform similar tests without given inlier-noise statistics, as shown in Fig.~\ref{Benchmarking-RA-NS-free}. TIVM$^\circ$ can still tolerate up to 70\% outliers (with errors lower than 2$^\circ$), shows higher accuracy than GNC-MinT and ADAPT-MinT and converges typically within 10 iterations (significantly faster than GNC-MinT and ADAPT-MinT).

\subsection{Point Cloud Registration}

Robust point cloud registration solves the optimal rigid transformation: $\boldsymbol{R}^{\star}\in SO(3)$ and $\boldsymbol{t}^{\star}\in \mathbb{R}^3$ that best align two 3D point clouds from putative correspondences $\{\boldsymbol{p}_i\leftrightarrow\boldsymbol{q}_i\}_{i=1}^N$ (usually established by feature matching techniques, e.g.~\cite{rusu2009fast}) which are corrupted by outliers. The residual error can be denoted with L2-norm as: $Re_i=\|\boldsymbol{R}^{\star}\boldsymbol{p}_i+\boldsymbol{t}^{\star}-\boldsymbol{q}_i\|_2$. 

\begin{figure*}[t]
\centering

\setlength\tabcolsep{1pt}
\addtolength{\tabcolsep}{1pt}

\begin{tabular}{ccc}

\begin{minipage}[t]{0.325\linewidth}
\centering
\includegraphics[width=1\linewidth]{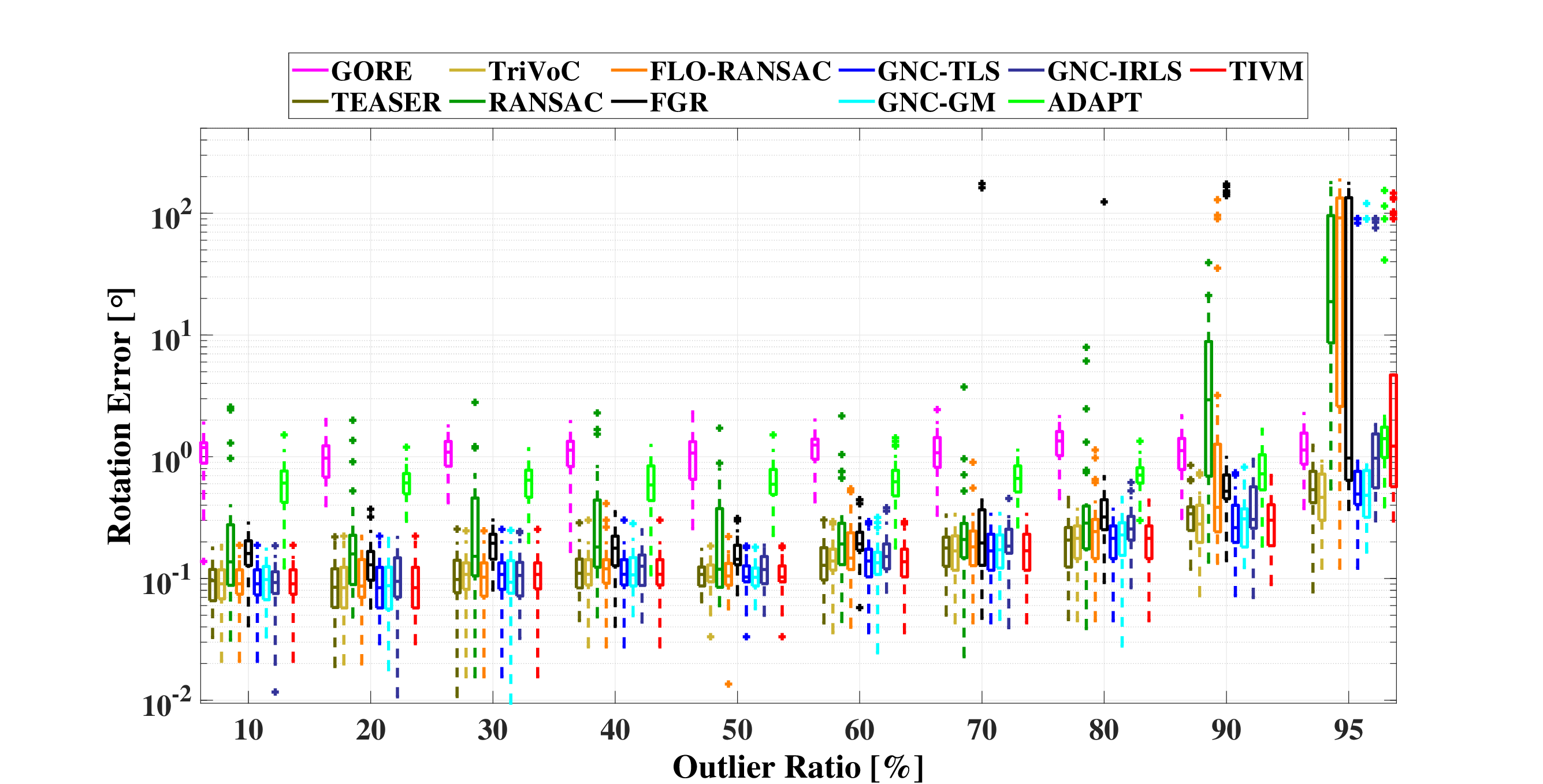}
\end{minipage}
&
\begin{minipage}[t]{0.325\linewidth}
\centering
\includegraphics[width=1\linewidth]{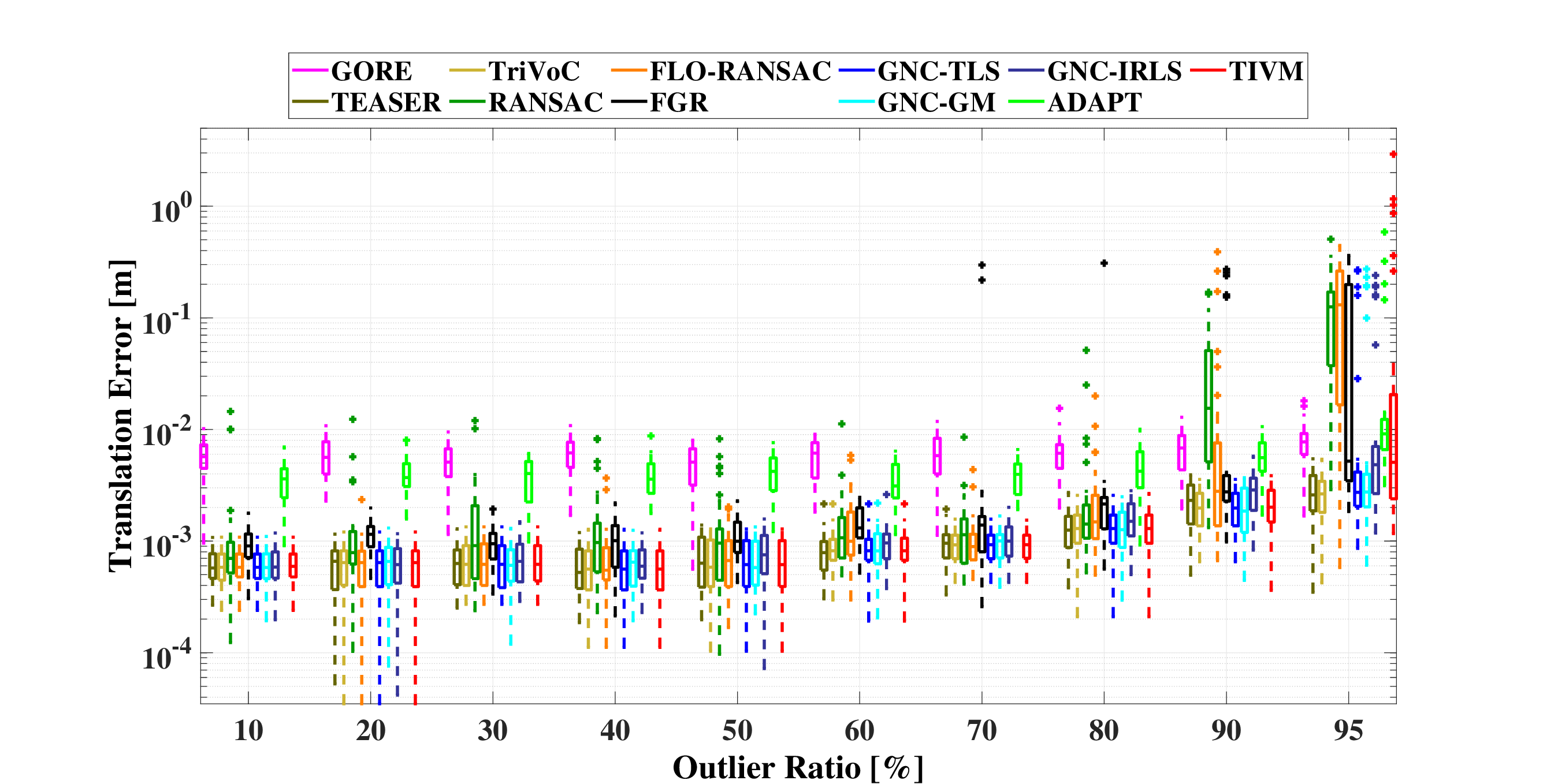}
\end{minipage}
&
\begin{minipage}[t]{0.31\linewidth}
\centering
\includegraphics[width=1\linewidth]{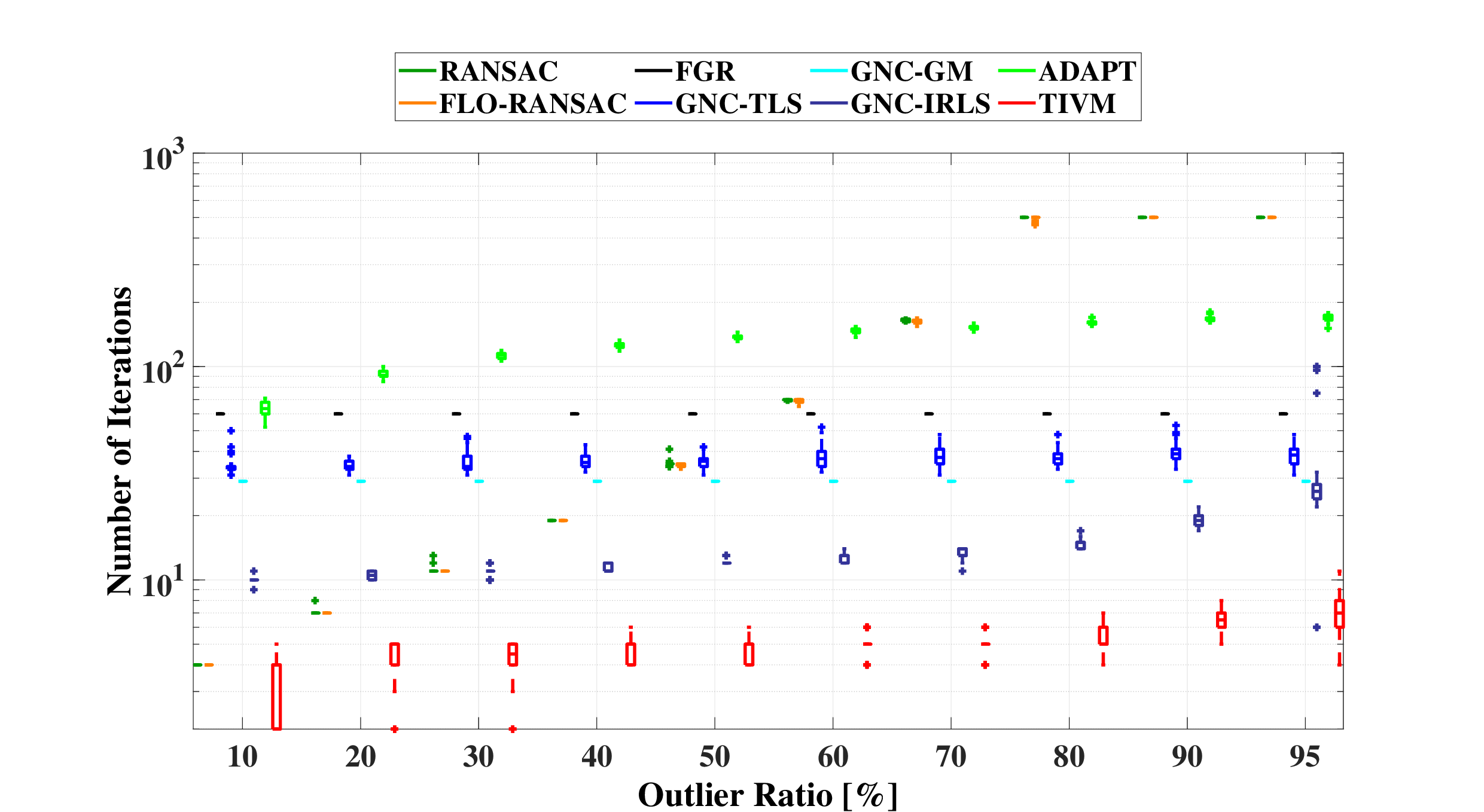}
\end{minipage}
\end{tabular}

\vspace{-1mm}
\caption{Benchmarking on robust point cloud registration.}
\label{Benchmarking-PCR}
\vspace{-3mm}
\end{figure*}

\textbf{Setup.} We adopt the \textit{bunny} from Stanford 3D Scan Repository~\cite{curless1996volumetric}. We first downsample it to $N=1000$ points as the source point cloud: $\{\boldsymbol{p}_i\}_{i=1}^N$ and then resize it to fit in a $1\times 1\times 1m$ cube. In each run, we generate a random ground-truth transformation: $\left(\boldsymbol{R}_{gt}\in SO(3),\boldsymbol{t}_{gt}\in\mathbb{R}^3\right)$ to transform $\{\boldsymbol{p}_i\}_{i=1}^N$, and then add random Gaussian noise with standard deviation $\sigma=0.01$ to obtain the target point cloud $\{\boldsymbol{q}_i\}_{i=1}^N$. Subsequently, a portion of the points in $\{\boldsymbol{q}_i\}_{i=1}^N$ are replaced with random 3D points near the point cloud surface to generate outliers. This setup is illustrated in Fig.~\ref{demo-PCR}. In terms of the specialized solvers, we use FGR~\cite{zhou2016fast}, GORE~\cite{bustos2017guaranteed}, TEASER~\cite{yang2020teaser} and TriVoC~\cite{sun2022trivoc}. We also include RANSAC~\cite{fischler1981random} and FLO-RANSAC~\cite{lebeda2012fixing} (both with 500 maximum iterations and $p=0.99$ confidence) as 2 state-of-the-art robust estimators for comparison. The inlier threshold is set to $\tau=5\sigma$. We adopt Horn's method~\cite{horn1987closed} as the minimal solver for RANSAC and FLO-RANSAC, and Arun's SVD approach~\cite{arun1987least} as the non-minimal solver. Benchmarking results over 30 Monte Carlo runs are reported in Fig.~\ref{Benchmarking-PCR}.

\begin{figure*}[t]
\centering

\setlength\tabcolsep{1pt}
\addtolength{\tabcolsep}{1pt}

\begin{tabular}{ccc}

\begin{minipage}[t]{0.25\linewidth}
\centering
\includegraphics[width=1\linewidth]{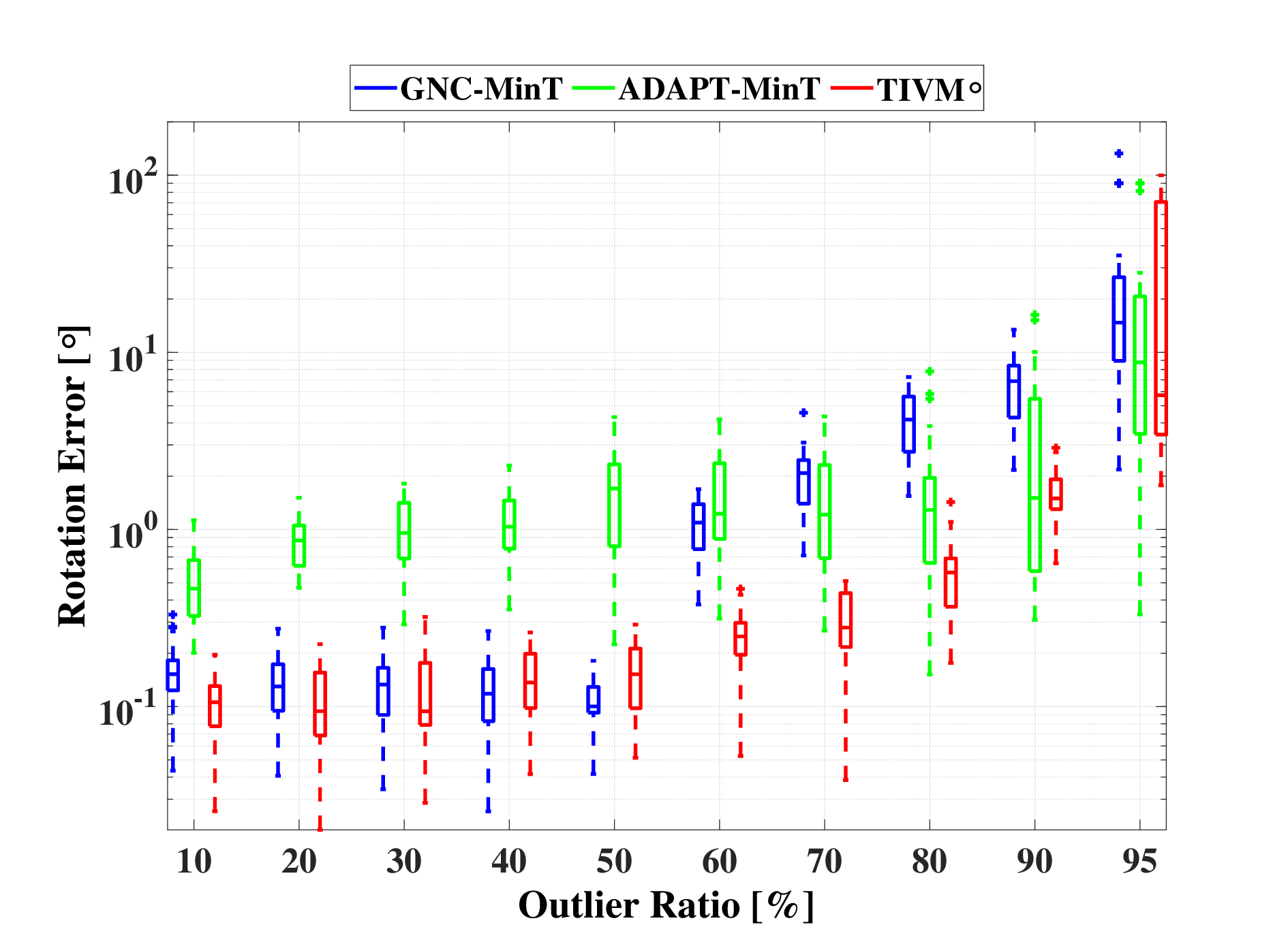}
\end{minipage}
&
\begin{minipage}[t]{0.25\linewidth}
\centering
\includegraphics[width=1\linewidth]{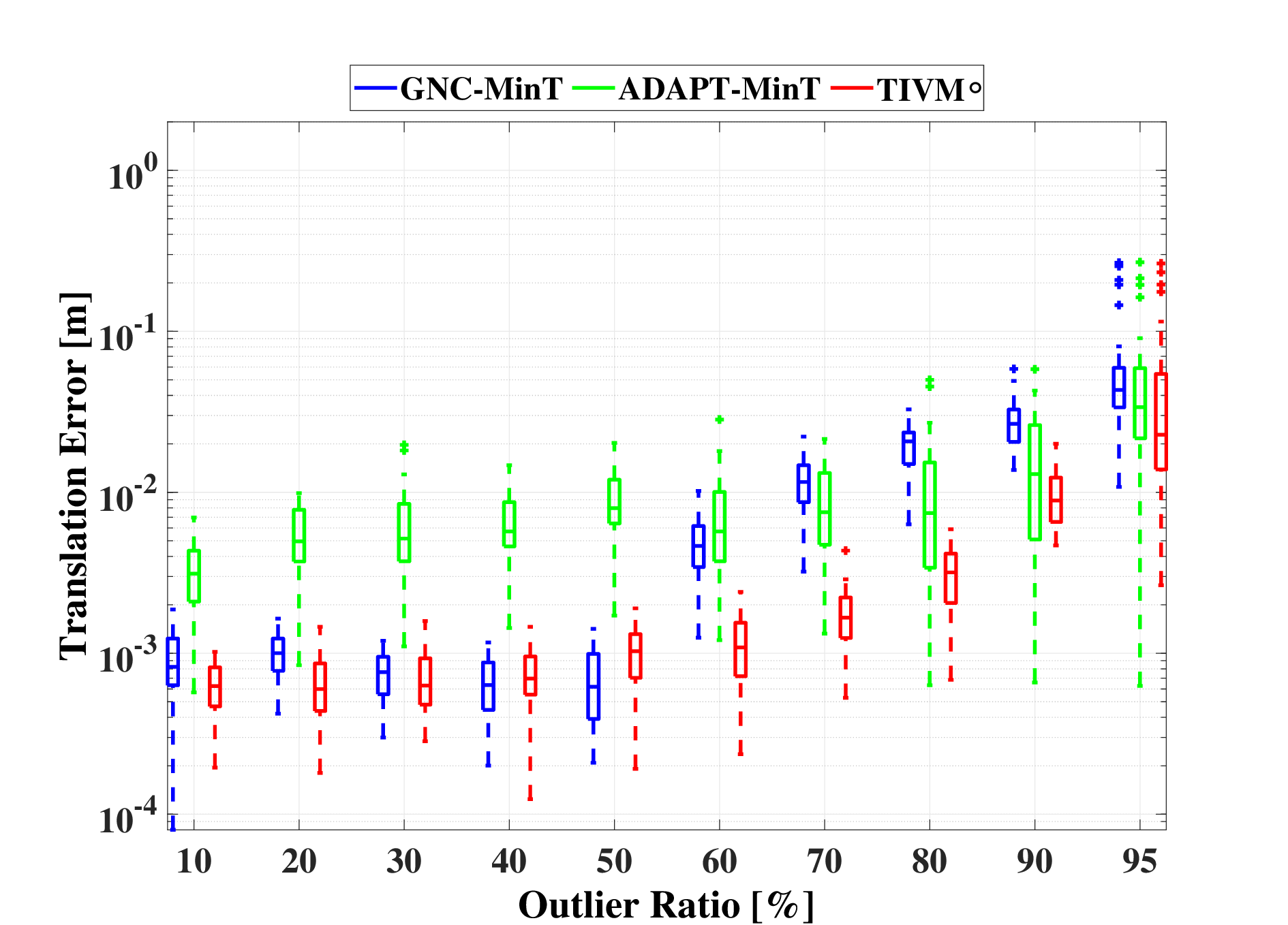}
\end{minipage}
&
\begin{minipage}[t]{0.25\linewidth}
\centering
\includegraphics[width=1\linewidth]{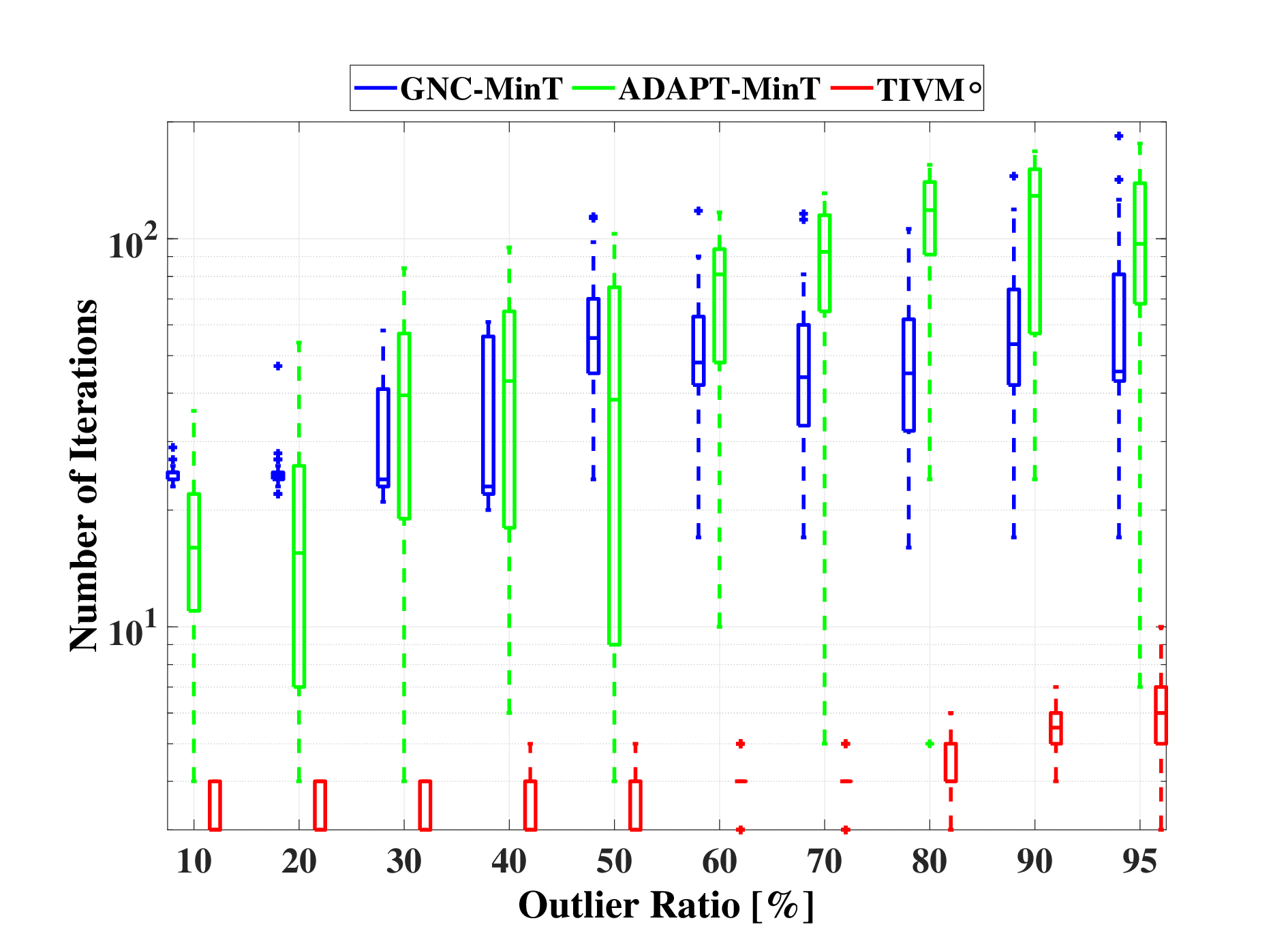}
\end{minipage}
\end{tabular}

\vspace{-1mm}
\caption{Benchmarking on robust point cloud registration without inlier-noise statistics.} 
\label{Benchmarking-PCR-NS-free}
\vspace{-3mm}
\end{figure*}

\begin{figure*}[t]
\centering

\setlength\tabcolsep{1pt}
\addtolength{\tabcolsep}{1pt}

\begin{tabular}{ccc}

\multicolumn{3}{c}{\footnotesize{(a) Known Inlier-Noise Statistics}} \\

\begin{minipage}[t]{0.25\linewidth}
\centering
\includegraphics[width=1\linewidth]{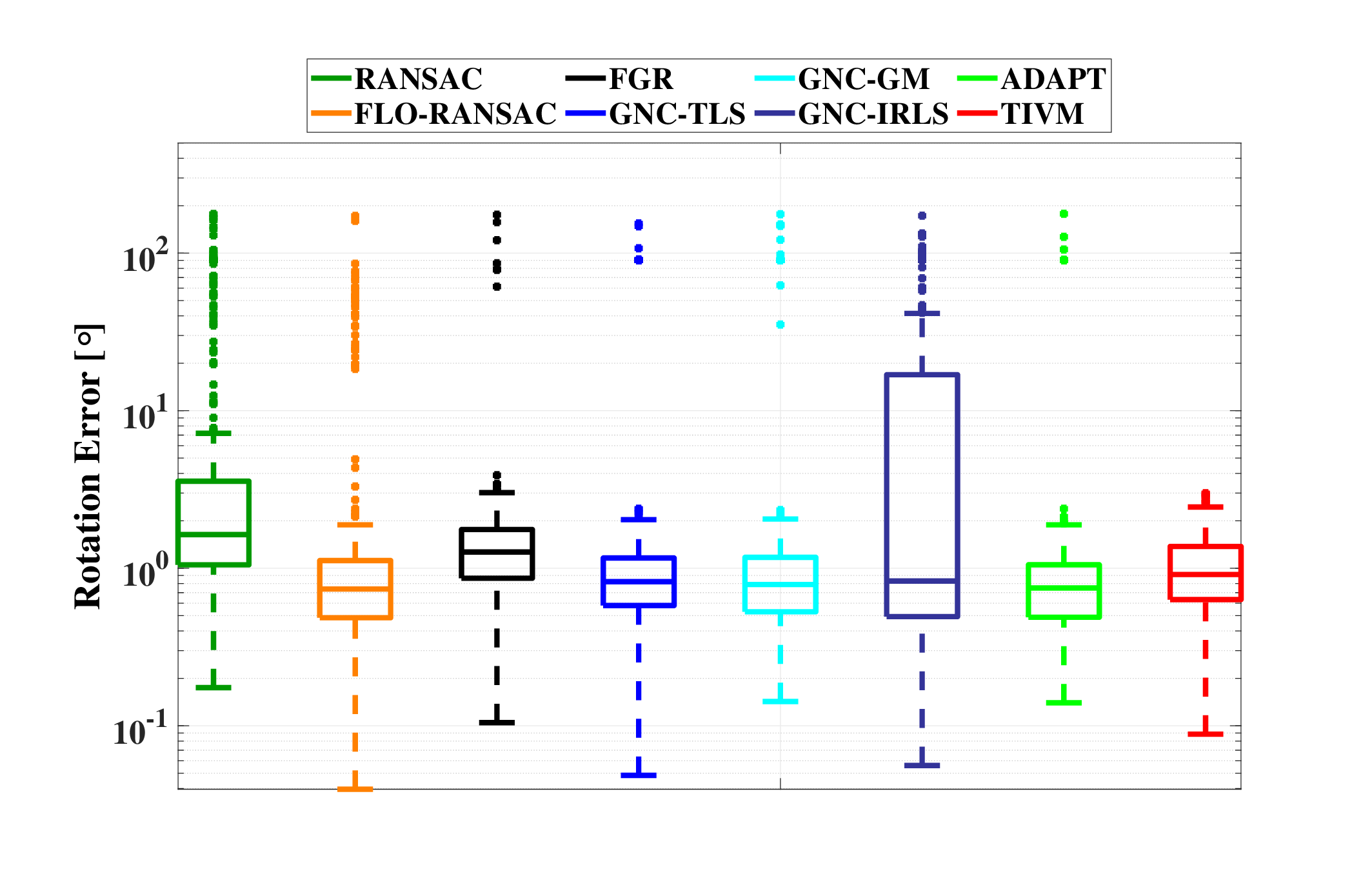}
\end{minipage}
&
\begin{minipage}[t]{0.25\linewidth}
\centering
\includegraphics[width=1\linewidth]{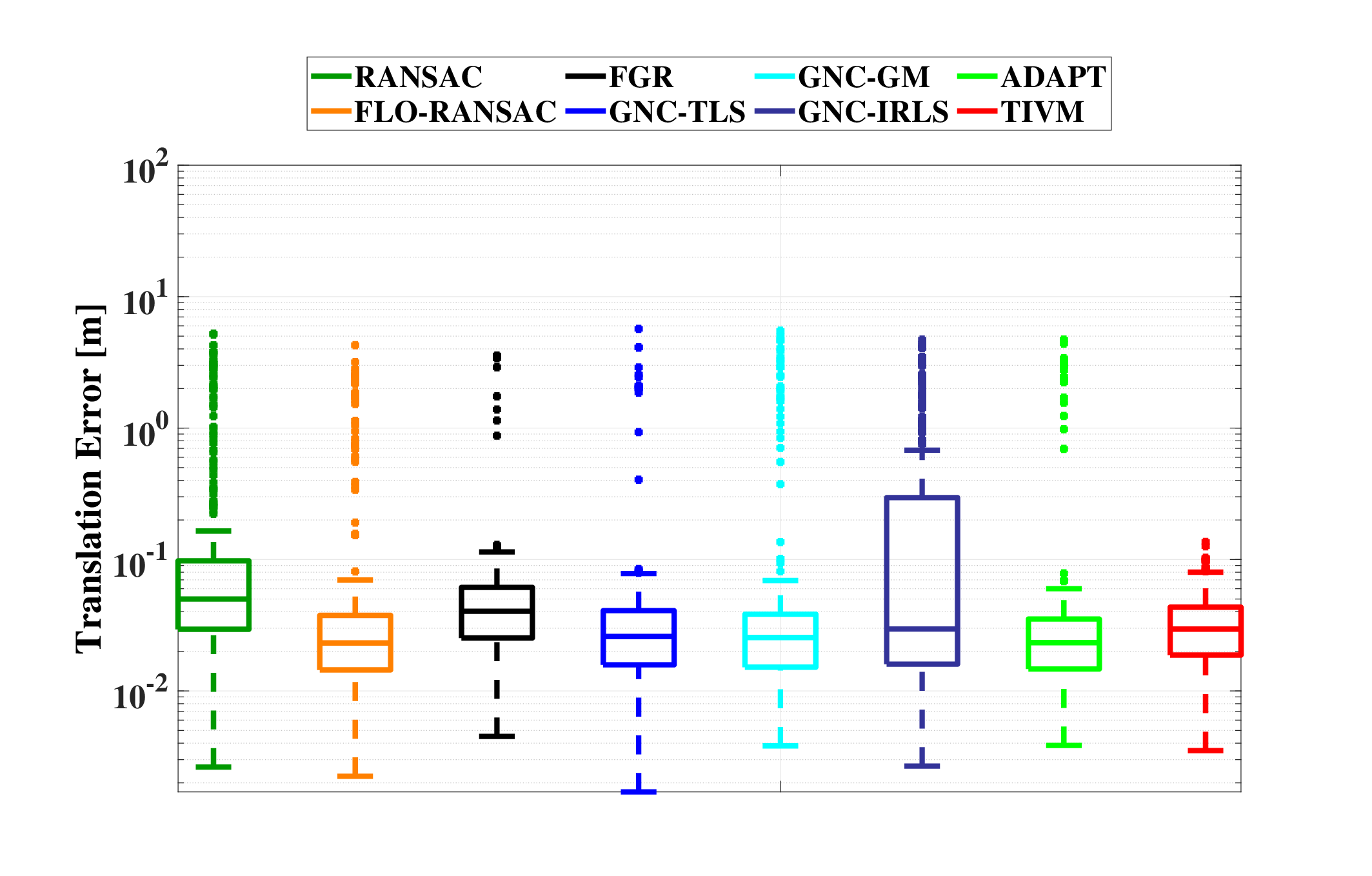}
\end{minipage}

&
\begin{minipage}[t]{0.25\linewidth}
\centering
\includegraphics[width=1\linewidth]{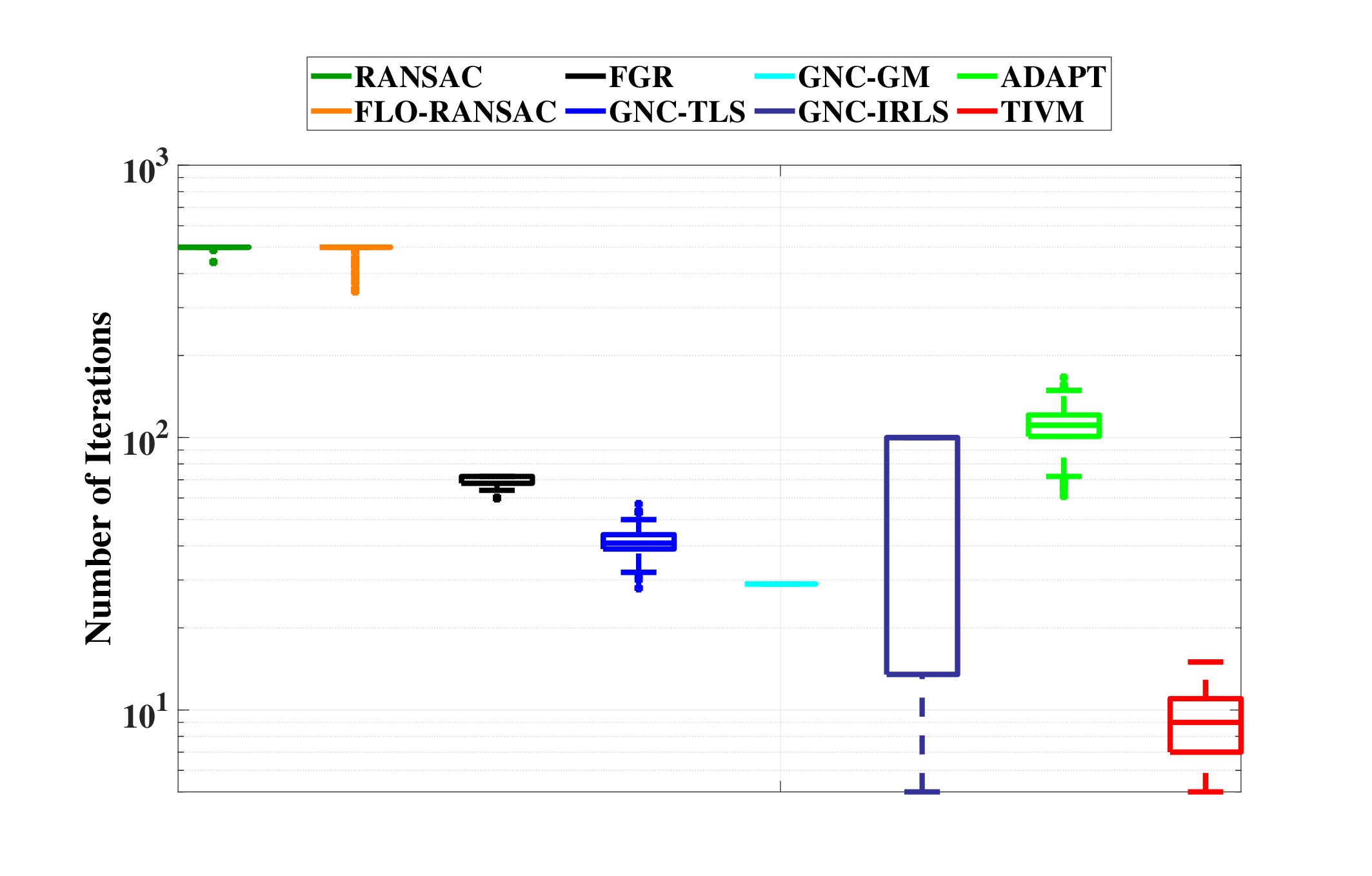}
\end{minipage}
\\

\multicolumn{3}{c}{\footnotesize{(b) Unknown Inlier-Noise Statistics}} \\

\begin{minipage}[t]{0.23\linewidth}
\centering
\includegraphics[width=1\linewidth]{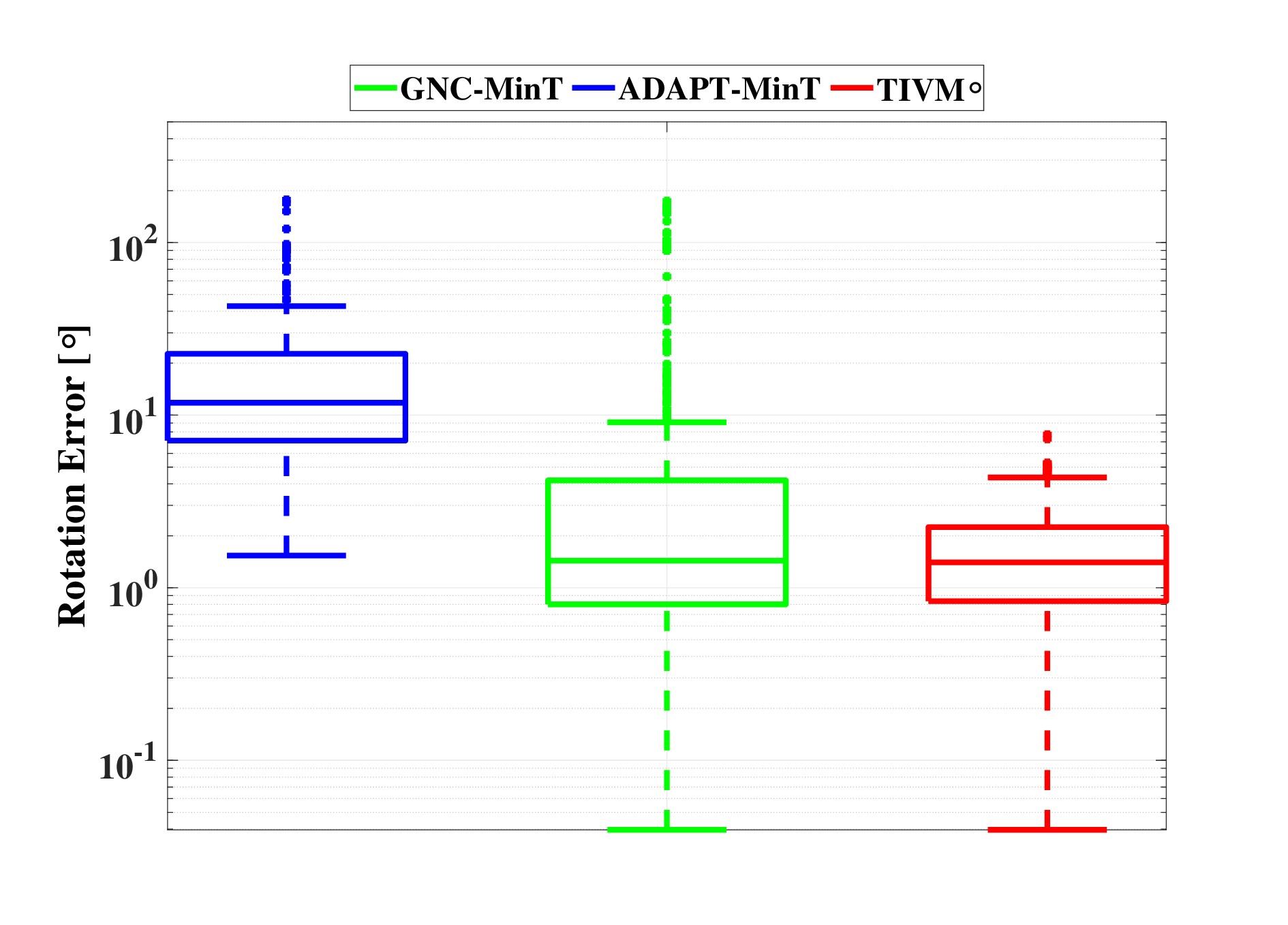}
\end{minipage}
&
\begin{minipage}[t]{0.23\linewidth}
\centering
\includegraphics[width=1\linewidth]{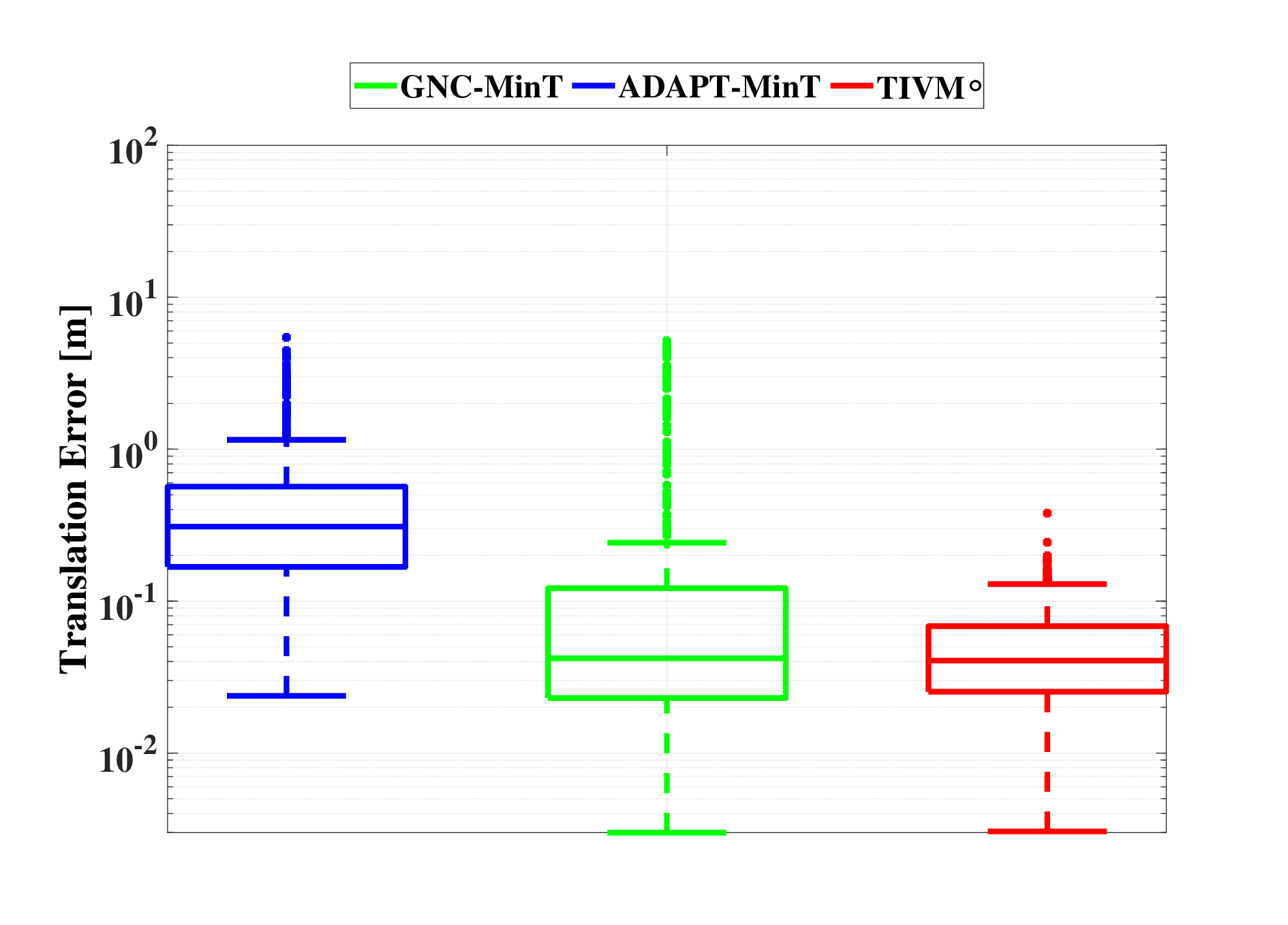}
\end{minipage}

&
\begin{minipage}[t]{0.23\linewidth}
\centering
\includegraphics[width=1\linewidth]{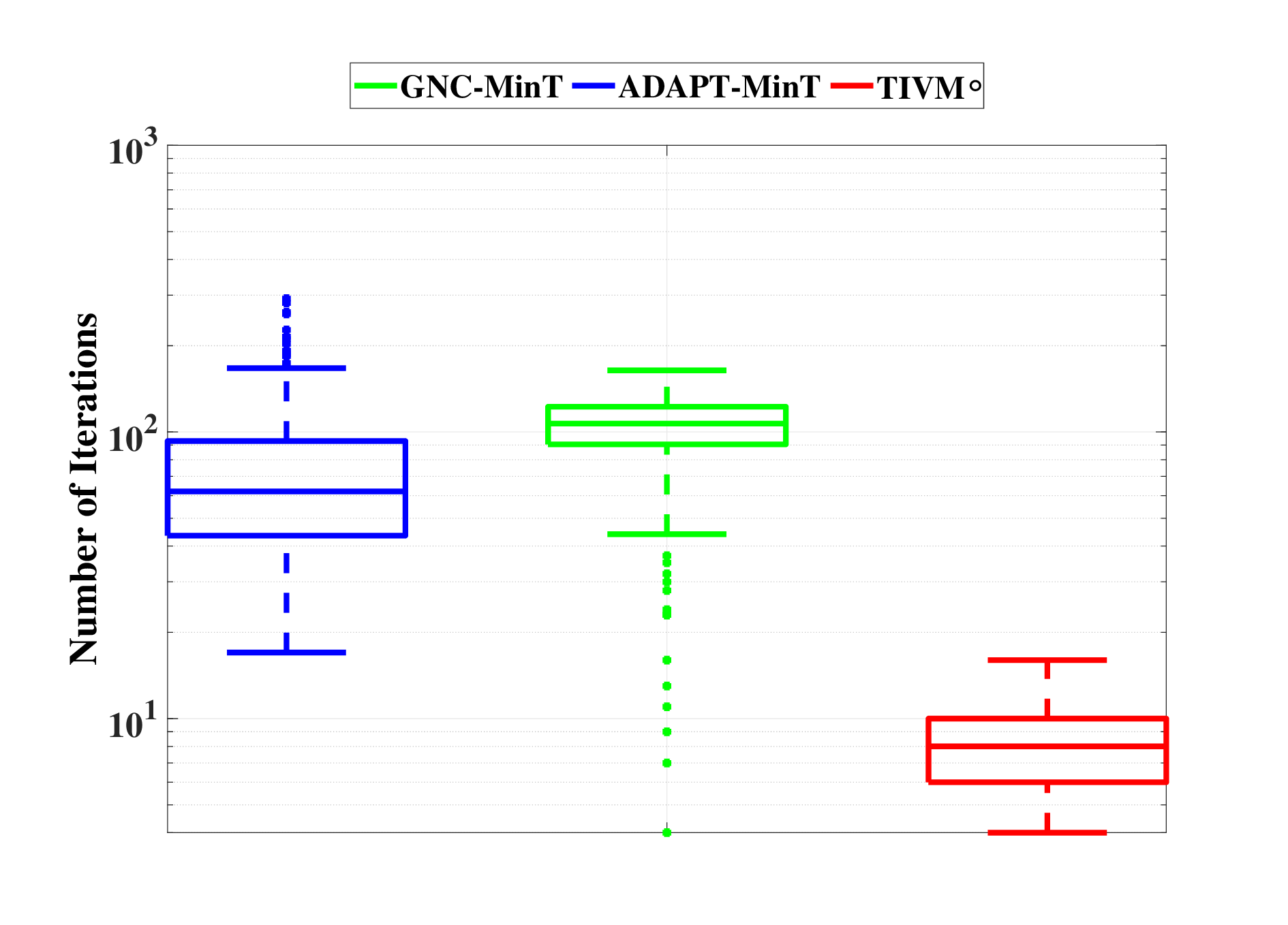}
\end{minipage}
\end{tabular}

\vspace{-1mm}
\caption{Quantitative results on scan stitching problems over 3DMatch dataset~\cite{zeng20173dmatch}.}
\label{3DMatch-PCR}
\vspace{-3mm}
\end{figure*}

\begin{figure*}[t]
\centering

\setlength\tabcolsep{2pt}
\addtolength{\tabcolsep}{3pt}

\begin{tabular}{cccc}

\multicolumn{1}{c}{\footnotesize{(a) 7-Scenes}} & \multicolumn{1}{c}{\footnotesize{(b) Hotel1}}  & \multicolumn{1}{c}{\footnotesize{(c) MIT Studyroom}} & \multicolumn{1}{c}{\footnotesize{(d) MIT Lab}} \\

\begin{minipage}[t]{0.2\linewidth}
\centering
\includegraphics[width=1\linewidth]{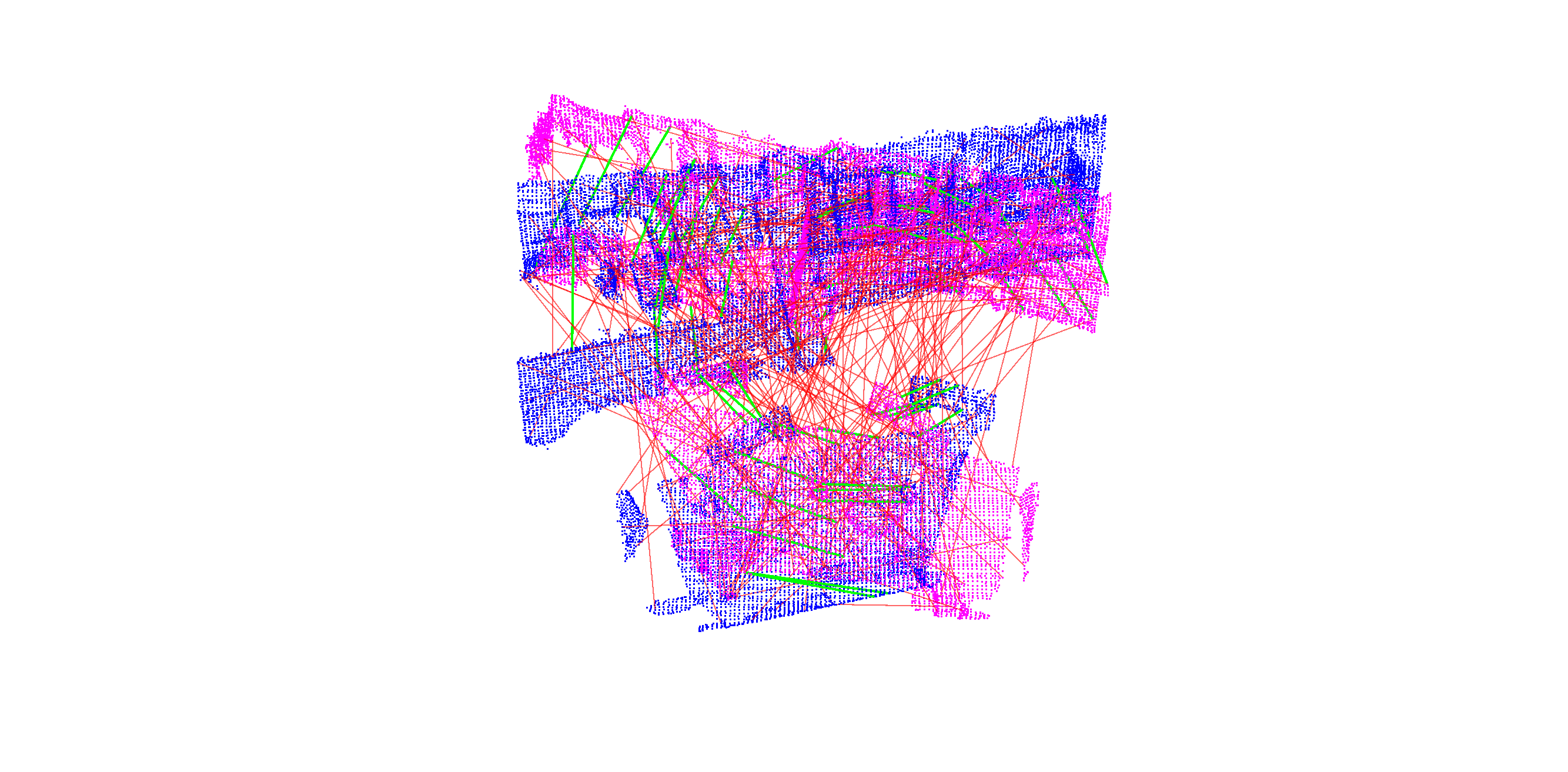}
\end{minipage}
&
\begin{minipage}[t]{0.2\linewidth}
\centering
\includegraphics[width=1\linewidth]{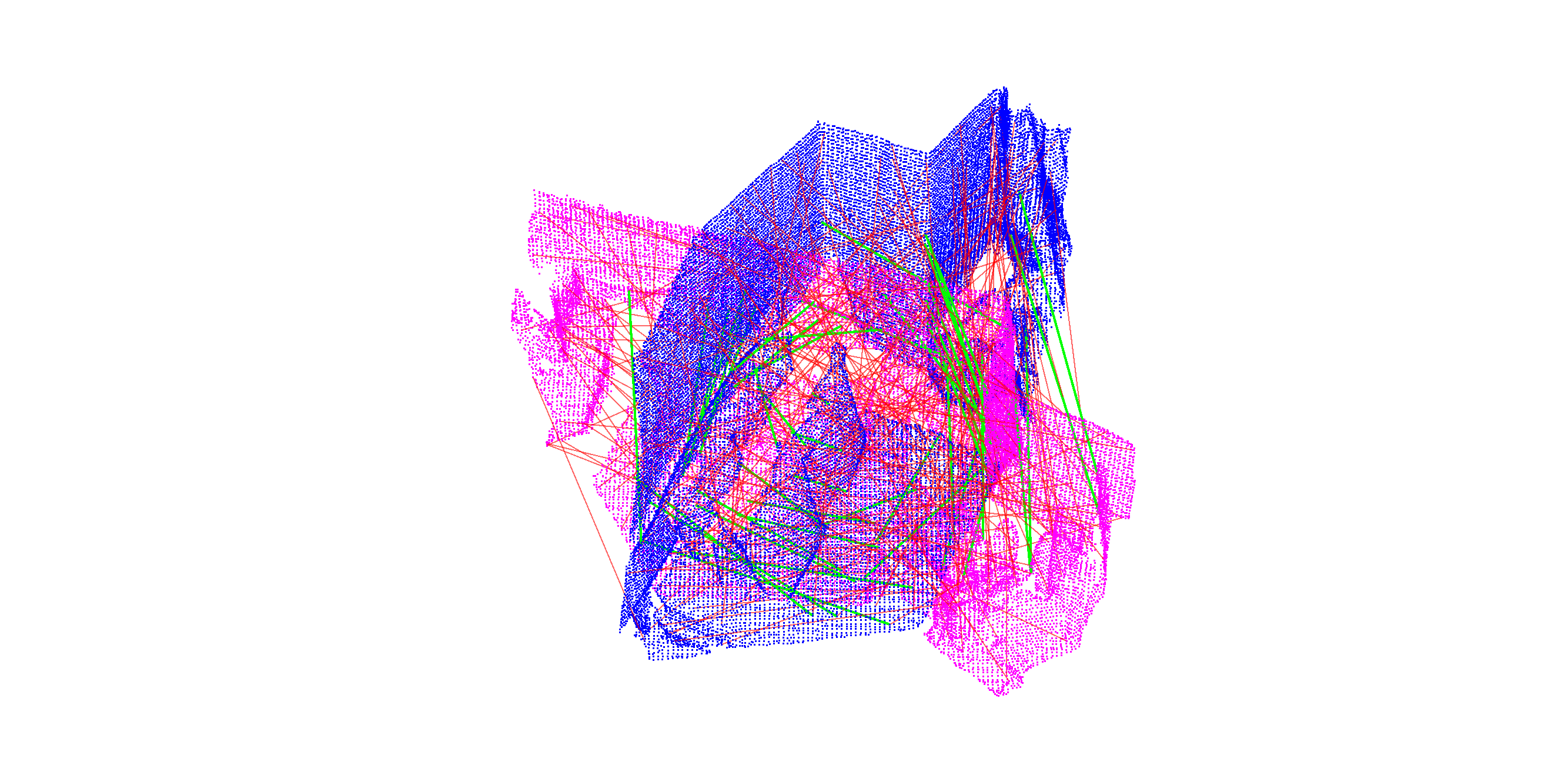}
\end{minipage}
&
\begin{minipage}[t]{0.2\linewidth}
\centering
\includegraphics[width=1\linewidth]{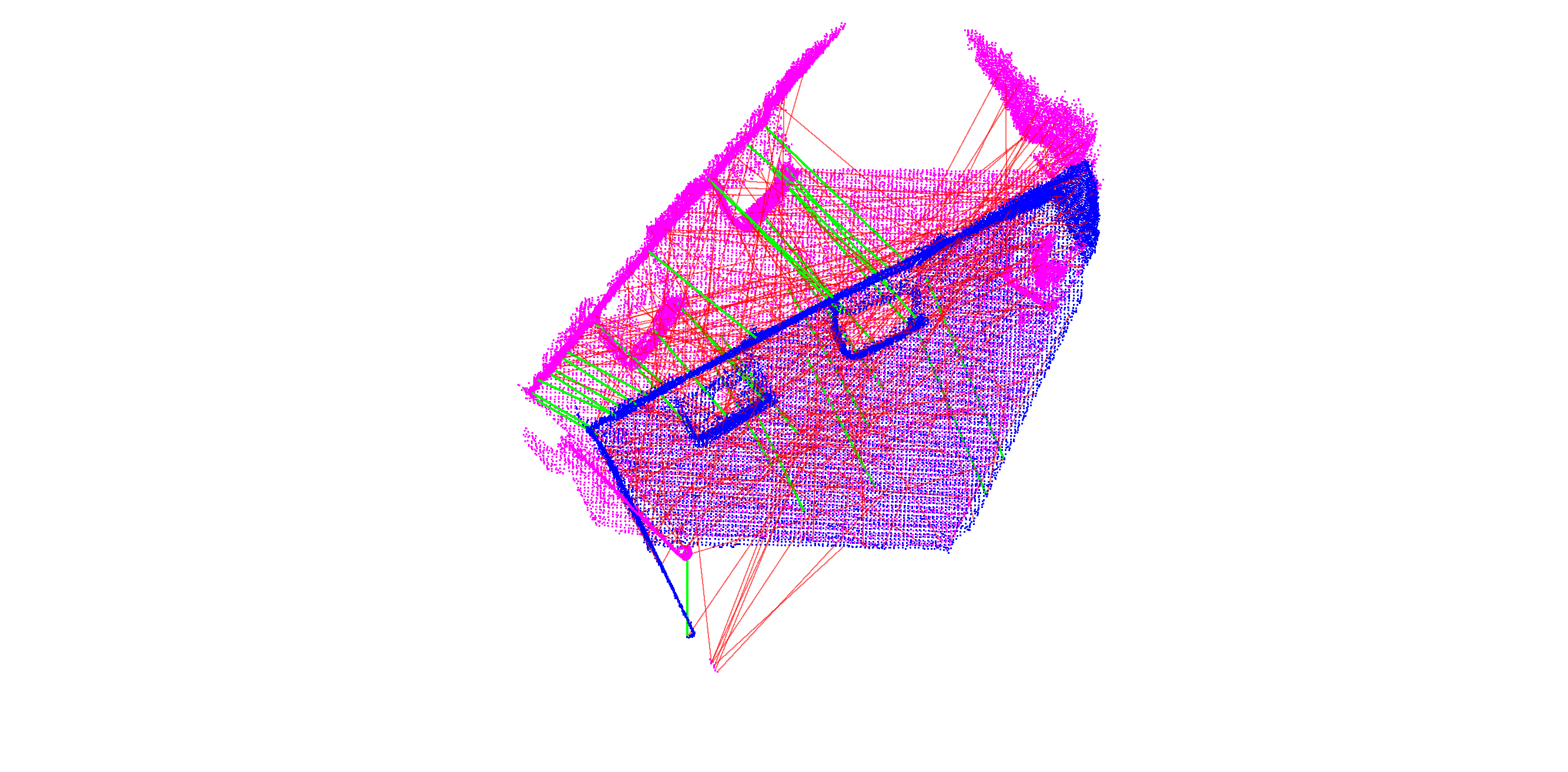}
\end{minipage}
&
\begin{minipage}[t]{0.2\linewidth}
\centering
\includegraphics[width=1\linewidth]{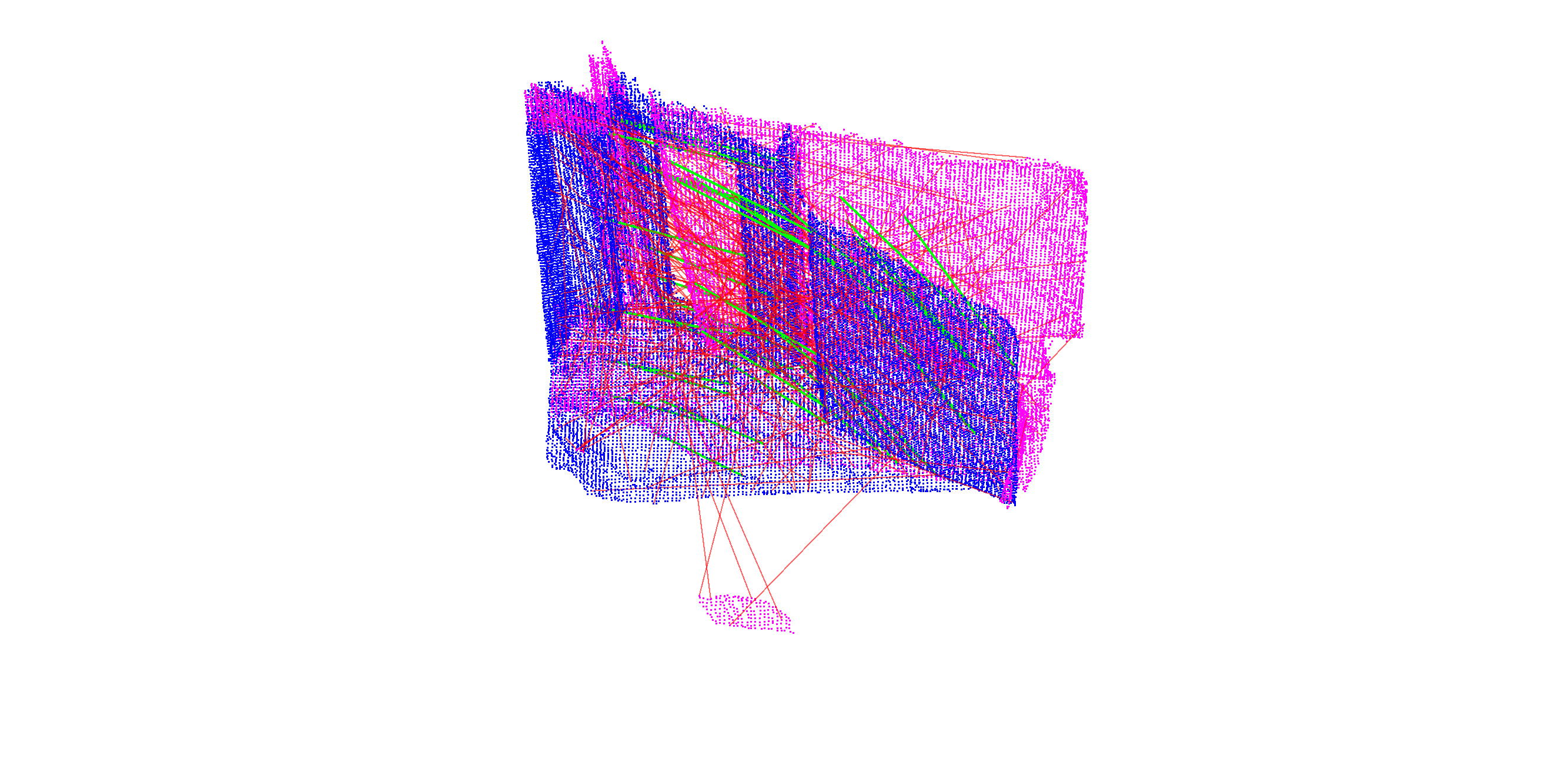}
\end{minipage}
\\
\begin{minipage}[t]{0.2\linewidth}
\centering
\includegraphics[width=1\linewidth]{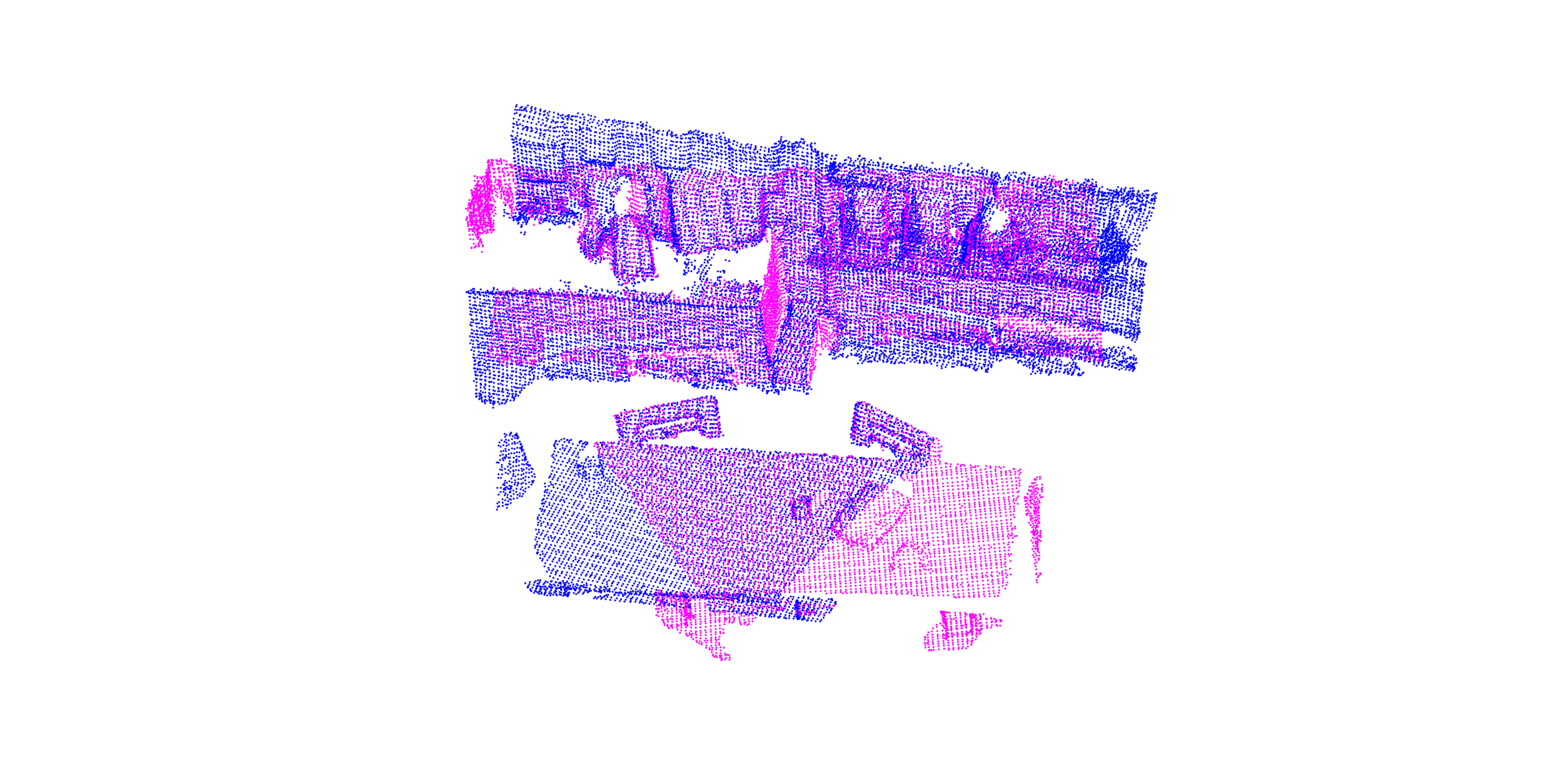}
\end{minipage}
&
\begin{minipage}[t]{0.2\linewidth}
\centering
\includegraphics[width=1\linewidth]{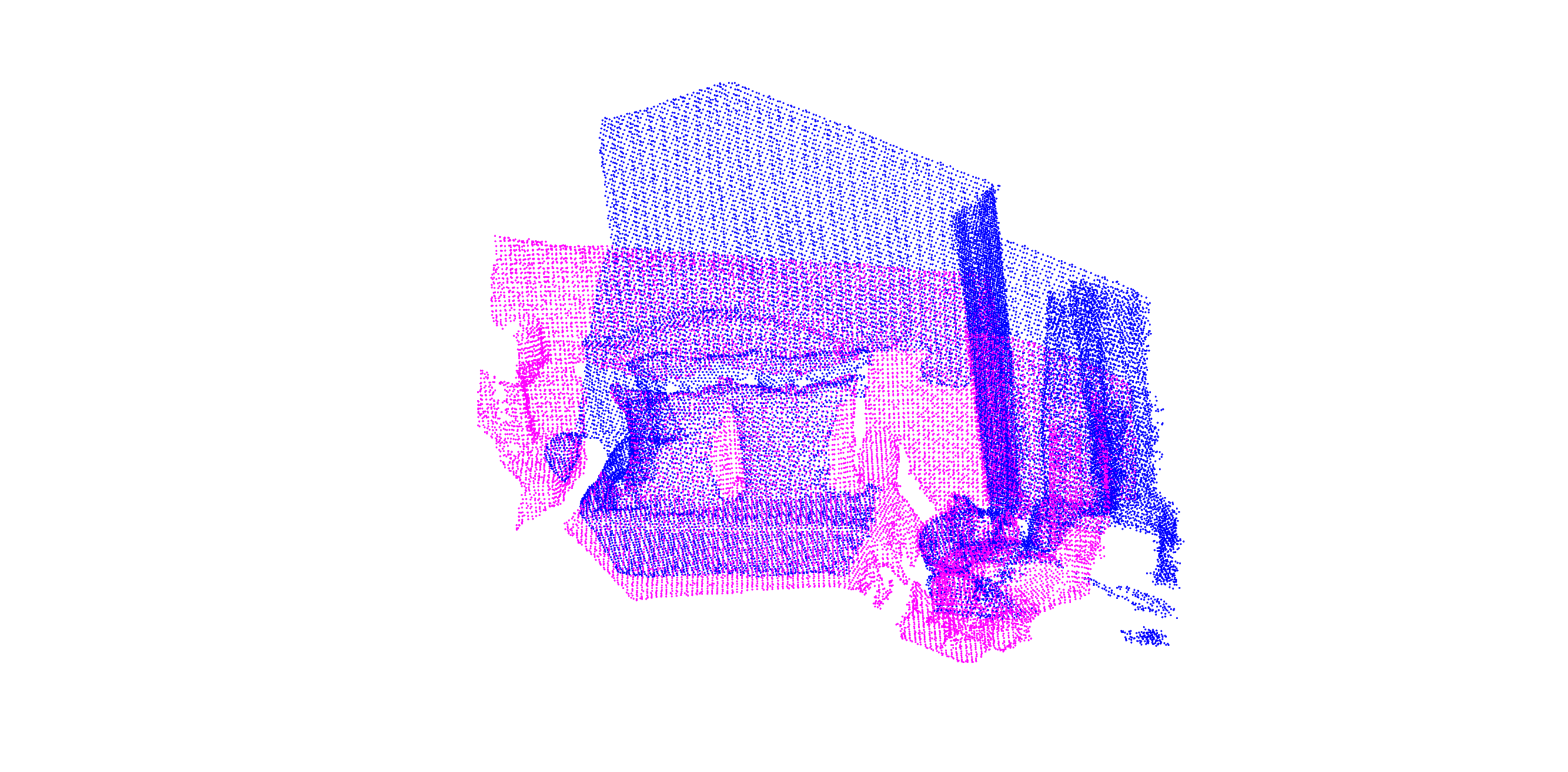}
\end{minipage}
&
\begin{minipage}[t]{0.2\linewidth}
\centering
\includegraphics[width=1\linewidth]{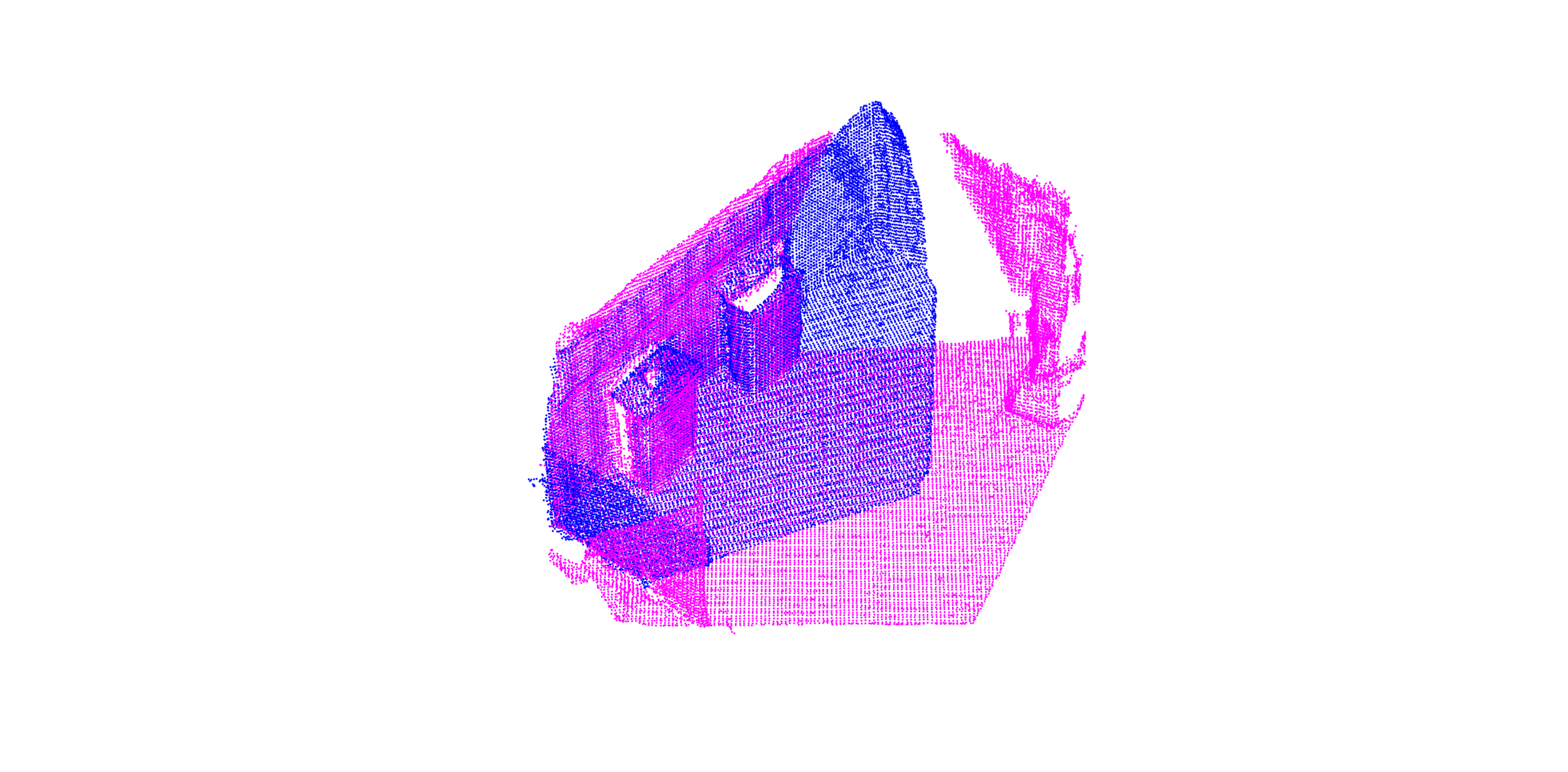}
\end{minipage}
&
\begin{minipage}[t]{0.2\linewidth}
\centering
\includegraphics[width=1\linewidth]{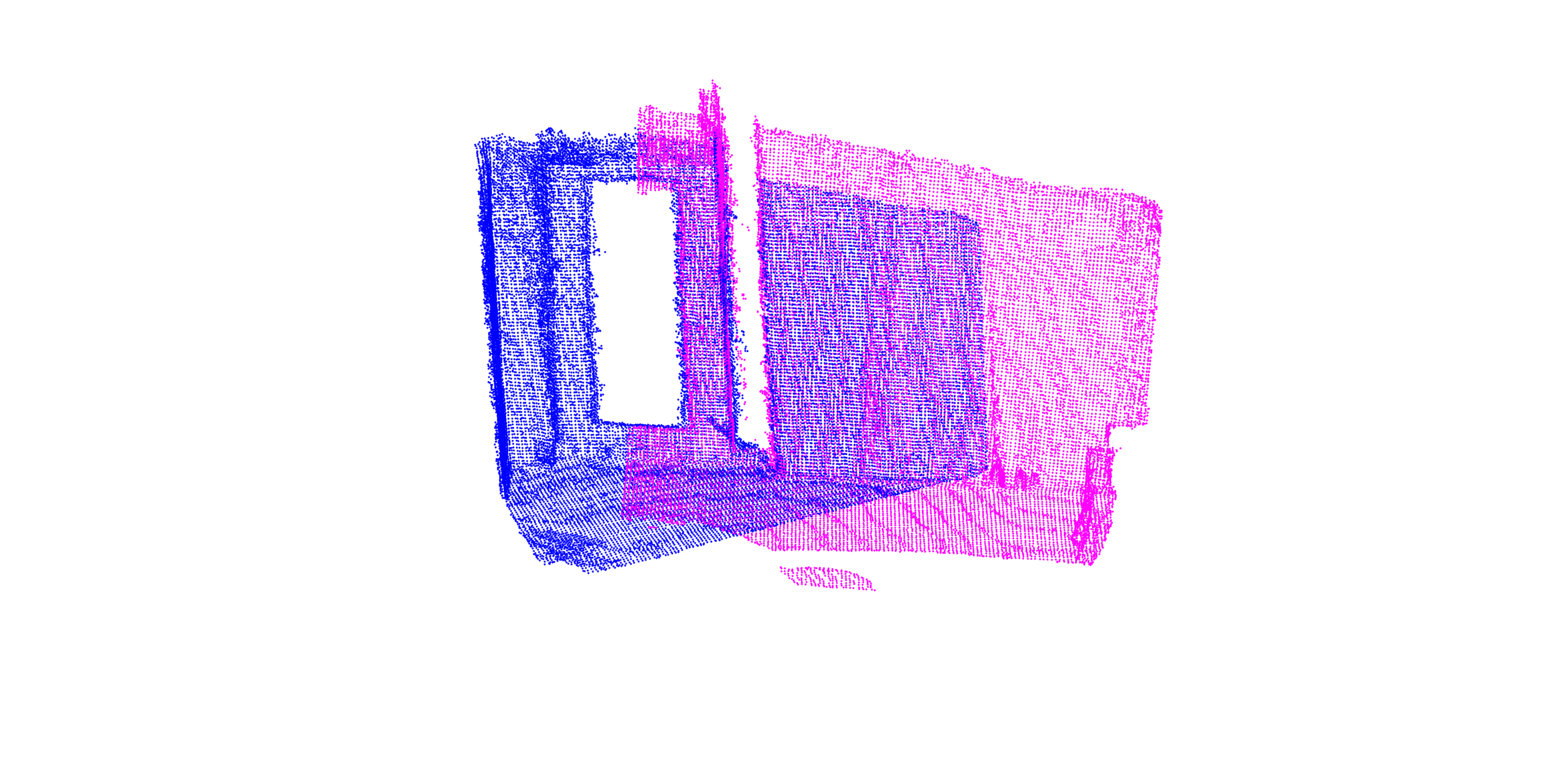}
\end{minipage}

\end{tabular}

\vspace{-2mm}
\caption{Qualitative examples of scan stitching over 3DMatch dataset~\cite{zeng20173dmatch}. \textbf{First Row:} Matched correspondences, where green and red lines denote inliers and outliers, respectively. \textbf{Second Row:} Scan stitching results using TIVM.} 
\label{3DMatch-PCR-qual}
\vspace{-5mm}
\end{figure*}

\begin{figure*}[t]
\centering

\setlength\tabcolsep{1pt}
\addtolength{\tabcolsep}{0pt}

\begin{tabular}{cccc}

\begin{minipage}[t]{0.25\linewidth}
\centering
\includegraphics[width=1\linewidth]{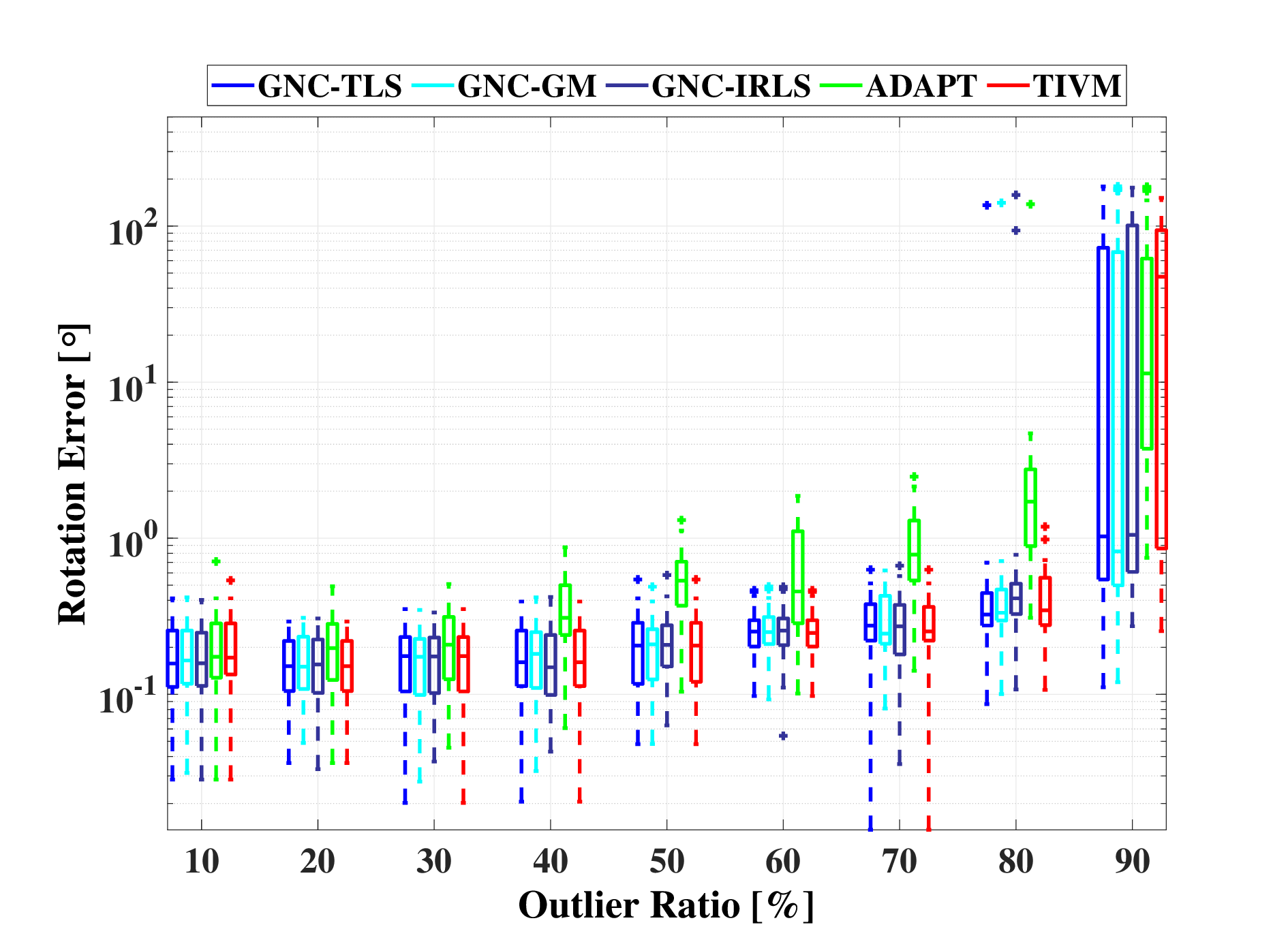}
\end{minipage}
&
\begin{minipage}[t]{0.25\linewidth}
\centering
\includegraphics[width=1\linewidth]{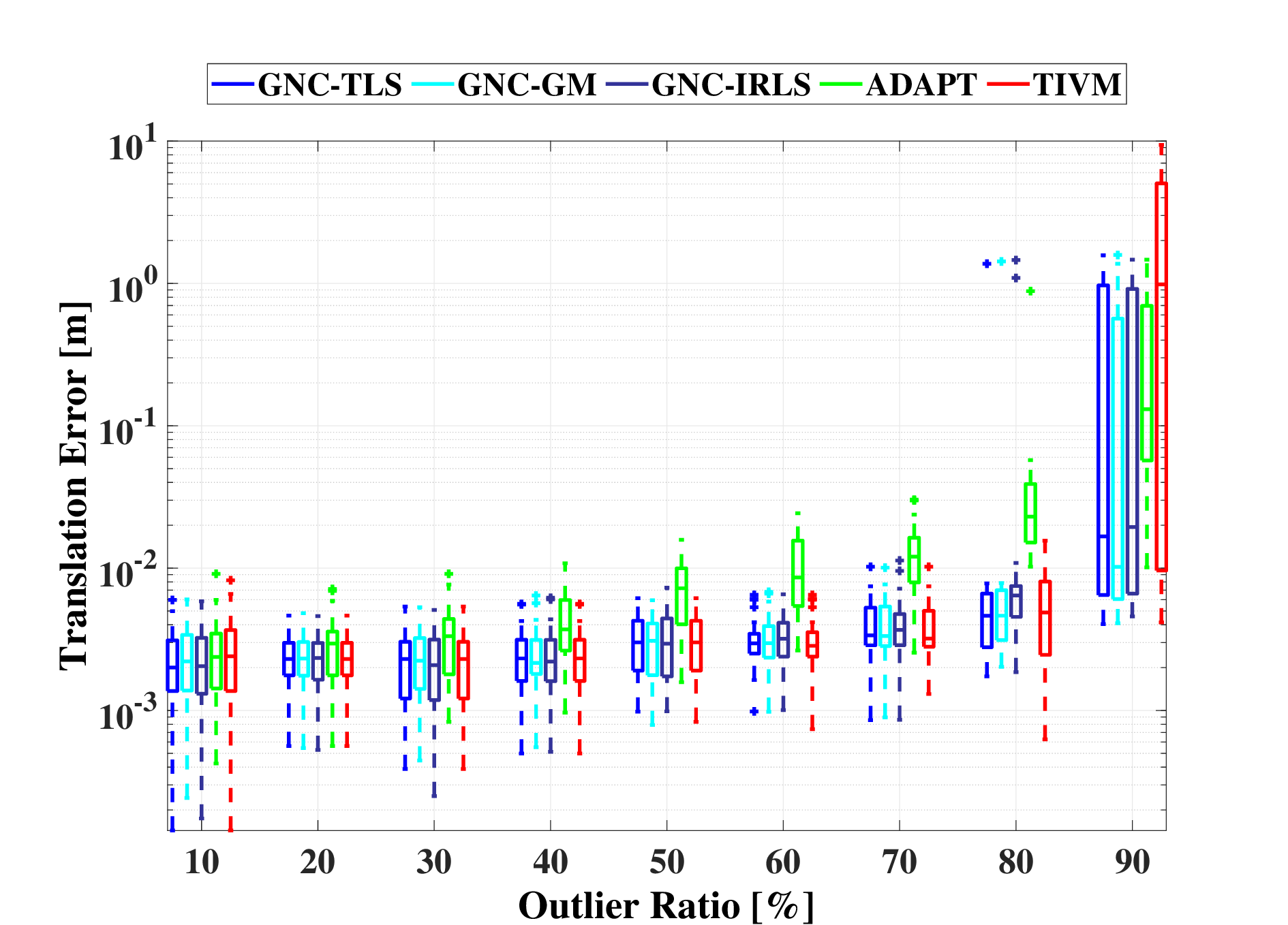}
\end{minipage}
&
\begin{minipage}[t]{0.25\linewidth}
\centering
\includegraphics[width=1\linewidth]{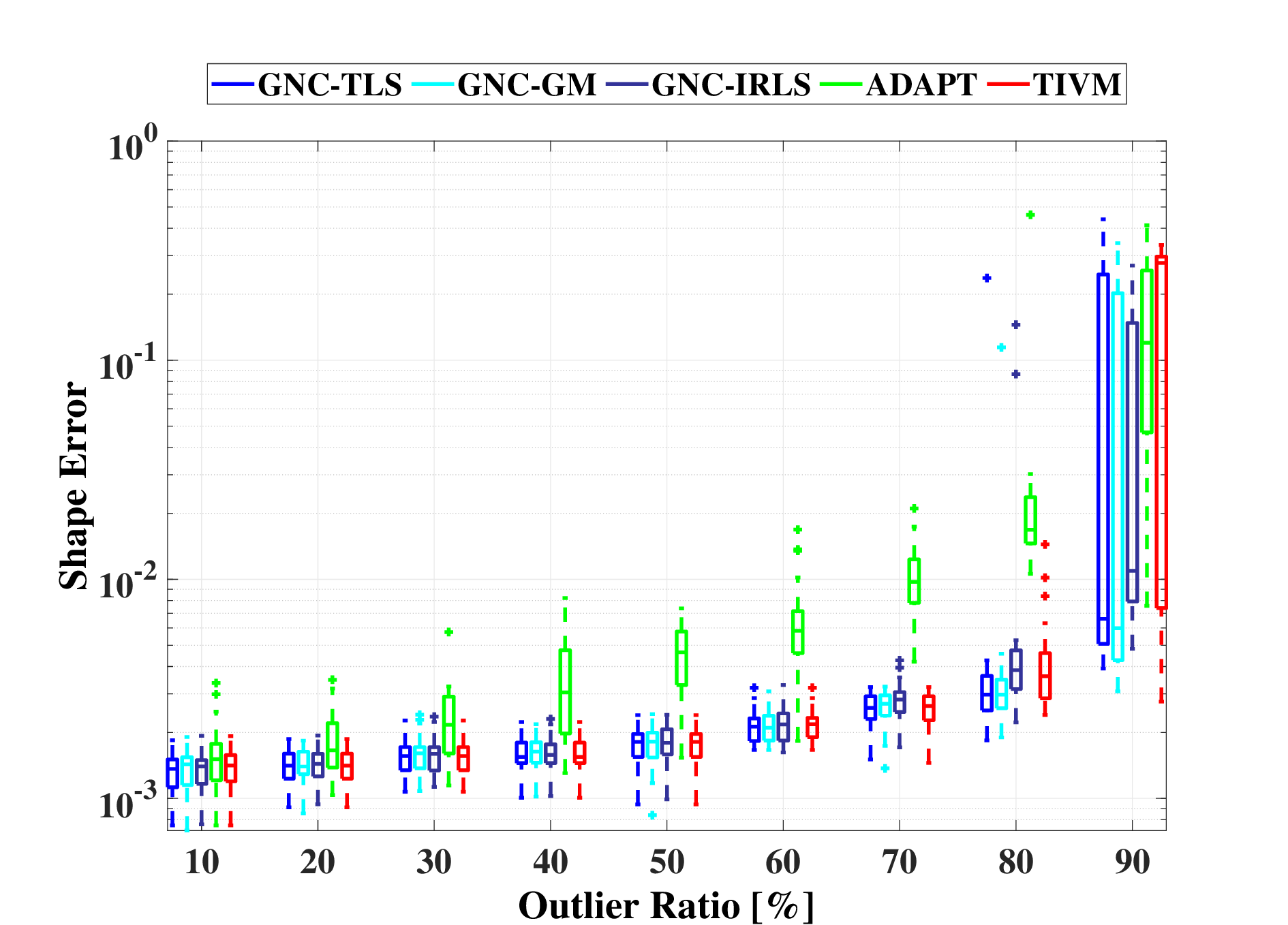}
\end{minipage}
&
\begin{minipage}[t]{0.25\linewidth}
\centering
\includegraphics[width=1\linewidth]{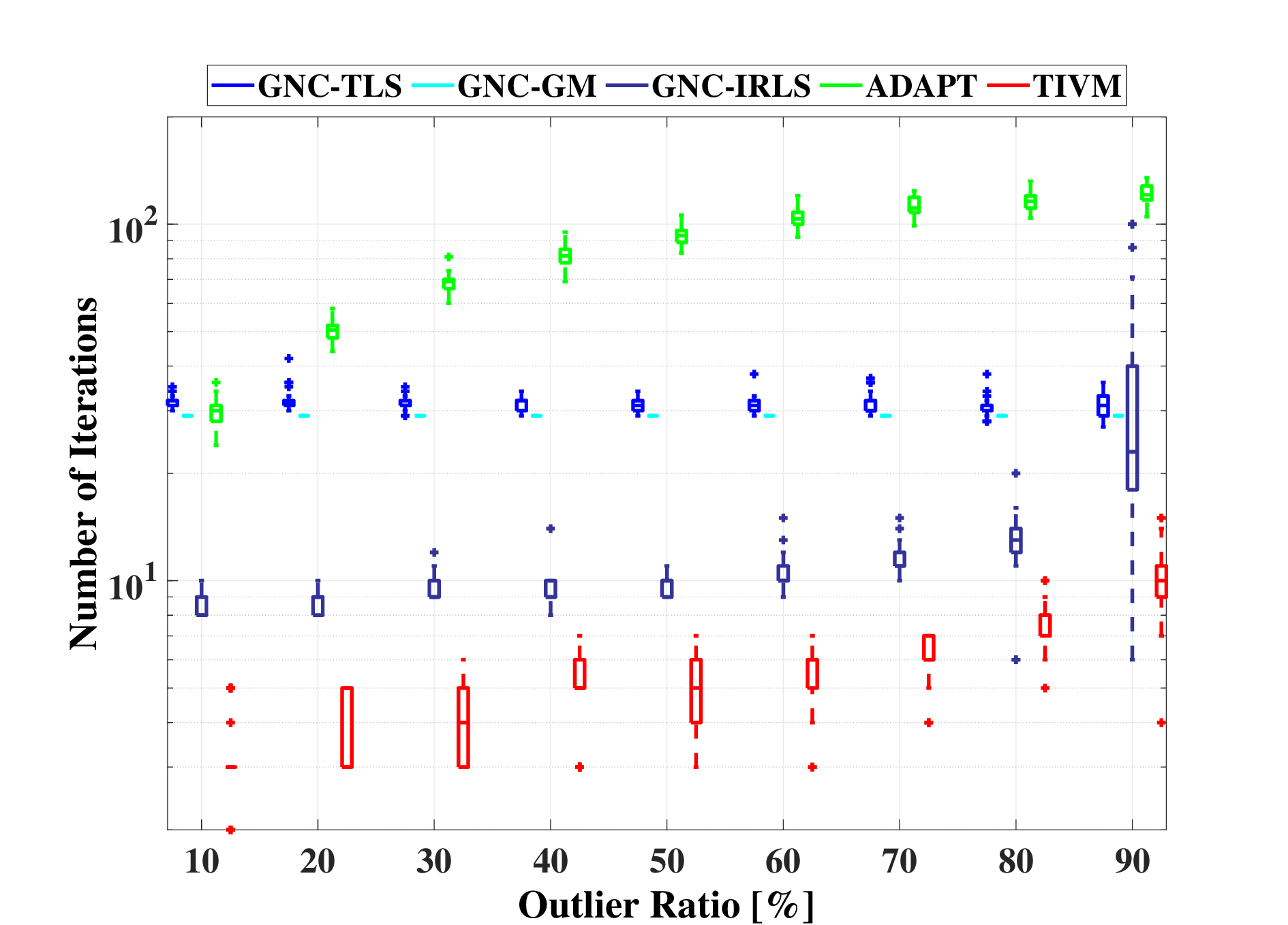}
\end{minipage}
\end{tabular}

\vspace{-1mm}
\caption{Benchmarking on robust category-level perception.}
\label{Benchmarking-CR}
\vspace{-1mm}
\end{figure*}

\begin{figure*}[t]
\centering

\setlength\tabcolsep{1pt}
\addtolength{\tabcolsep}{0pt}

\begin{tabular}{cccc}

\begin{minipage}[t]{0.25\linewidth}
\centering
\includegraphics[width=1\linewidth]{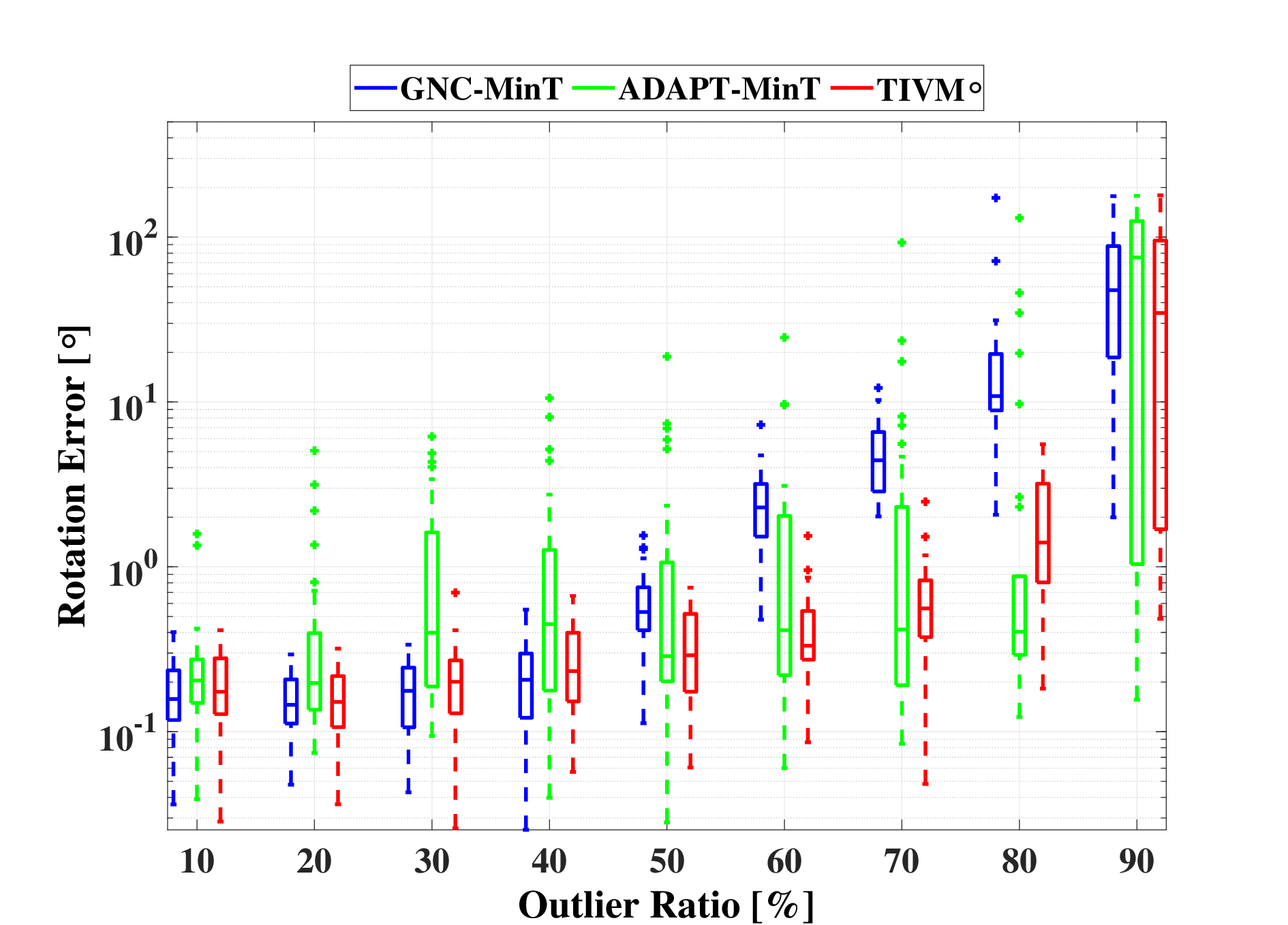}
\end{minipage}
&
\begin{minipage}[t]{0.25\linewidth}
\centering
\includegraphics[width=1\linewidth]{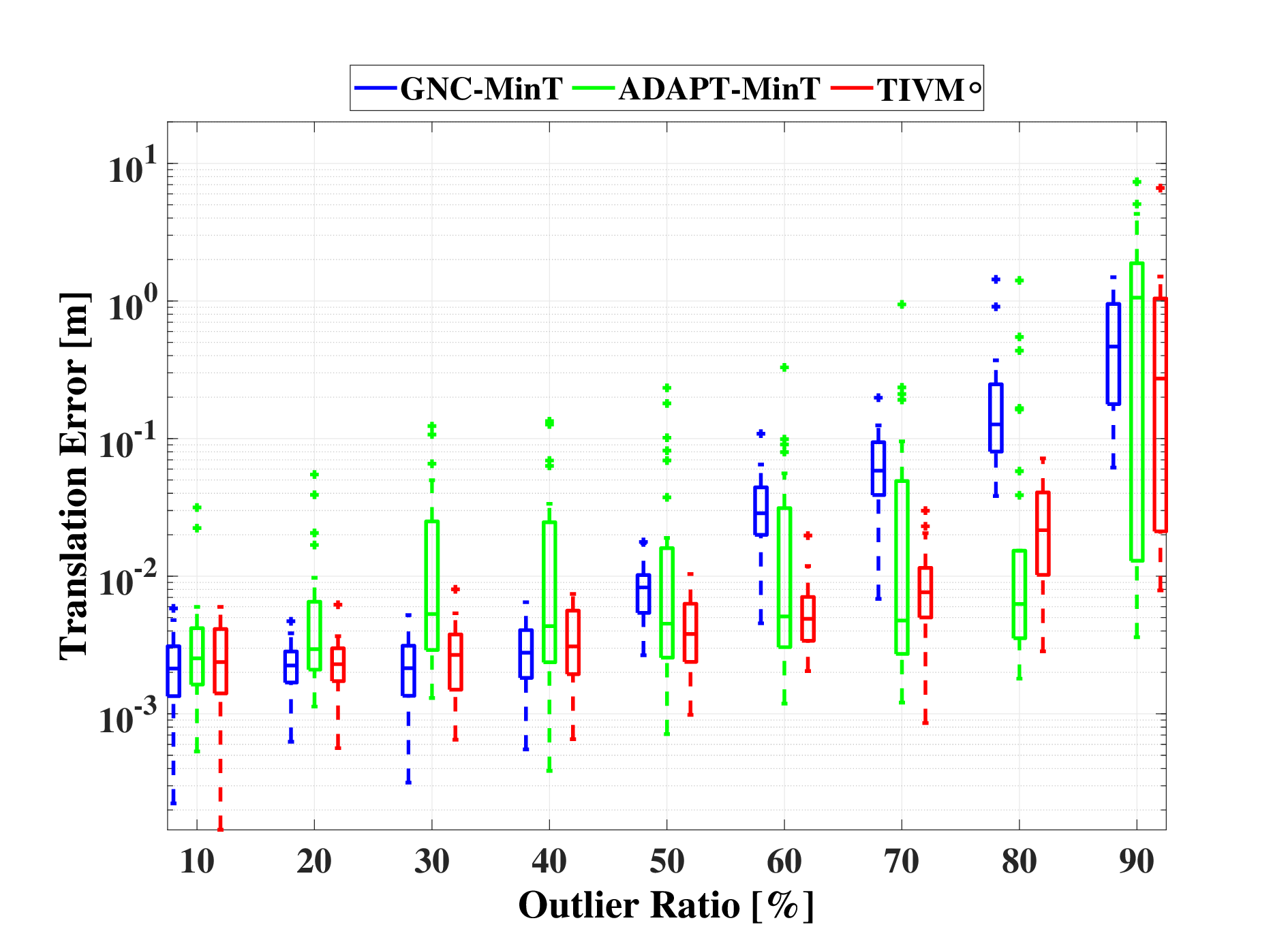}
\end{minipage}
&
\begin{minipage}[t]{0.25\linewidth}
\centering
\includegraphics[width=1\linewidth]{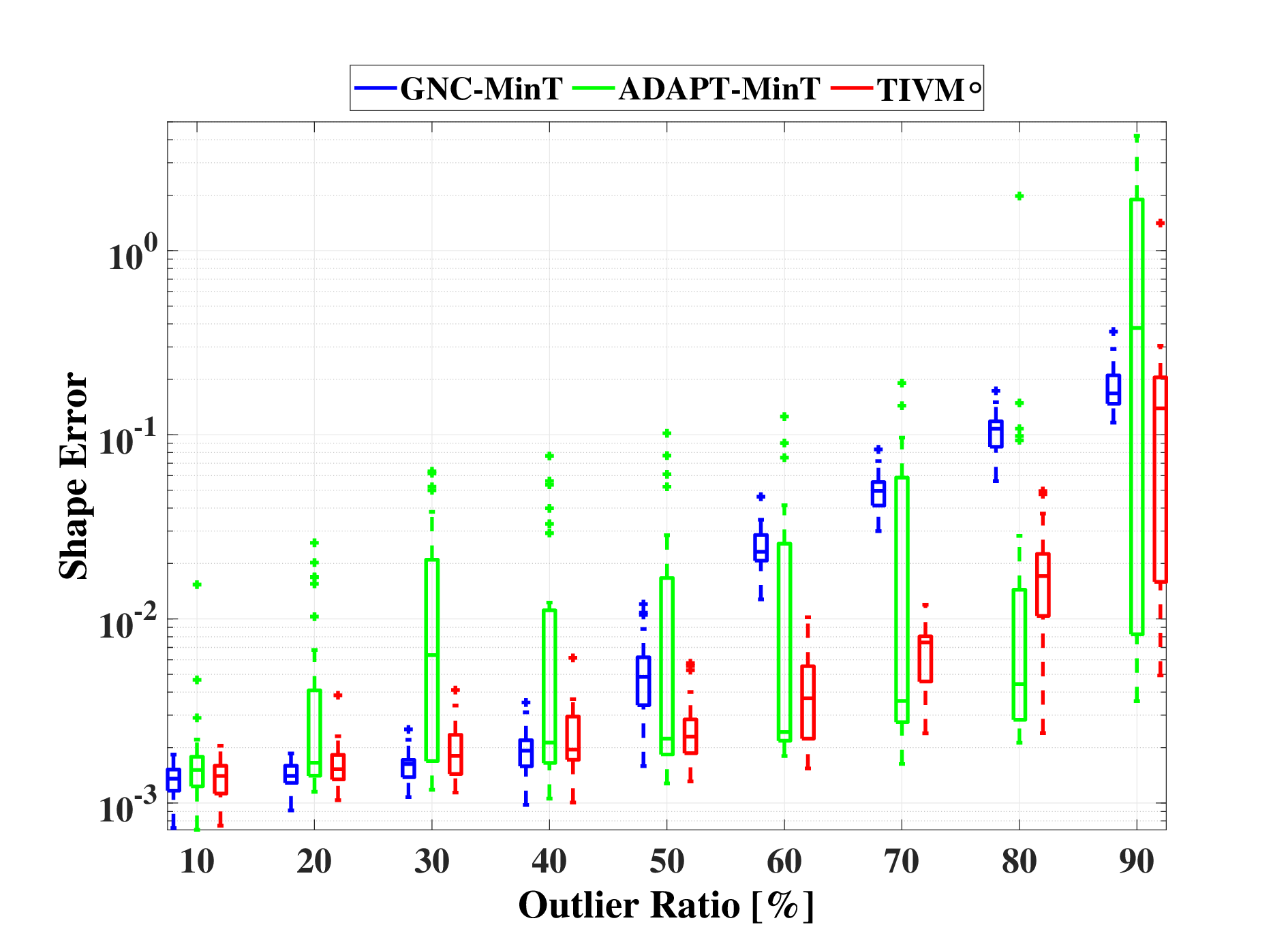}
\end{minipage}
&
\begin{minipage}[t]{0.25\linewidth}
\centering
\includegraphics[width=1\linewidth]{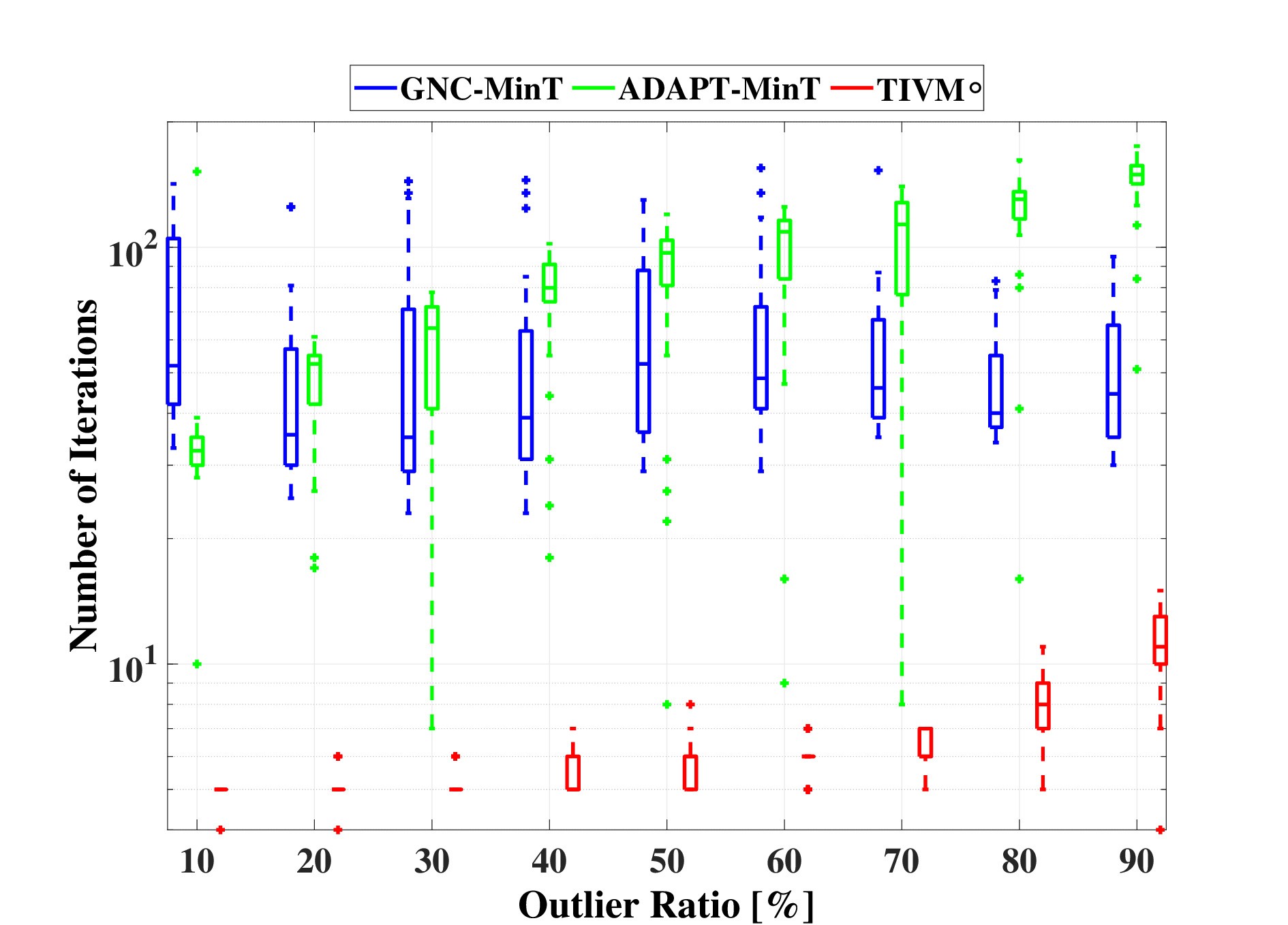}
\end{minipage}
\end{tabular}

\vspace{-1mm}
\caption{Benchmarking on robust category-level perception without inlier-noise statistics.}
\label{Benchmarking-CR-NS-free}
\vspace{-5mm}
\end{figure*}

\begin{figure}[t]
\centering

\setlength\tabcolsep{1pt}
\addtolength{\tabcolsep}{1pt}

\begin{tabular}{c}

\begin{minipage}[t]{0.6\linewidth}
\centering
\includegraphics[width=1\linewidth]{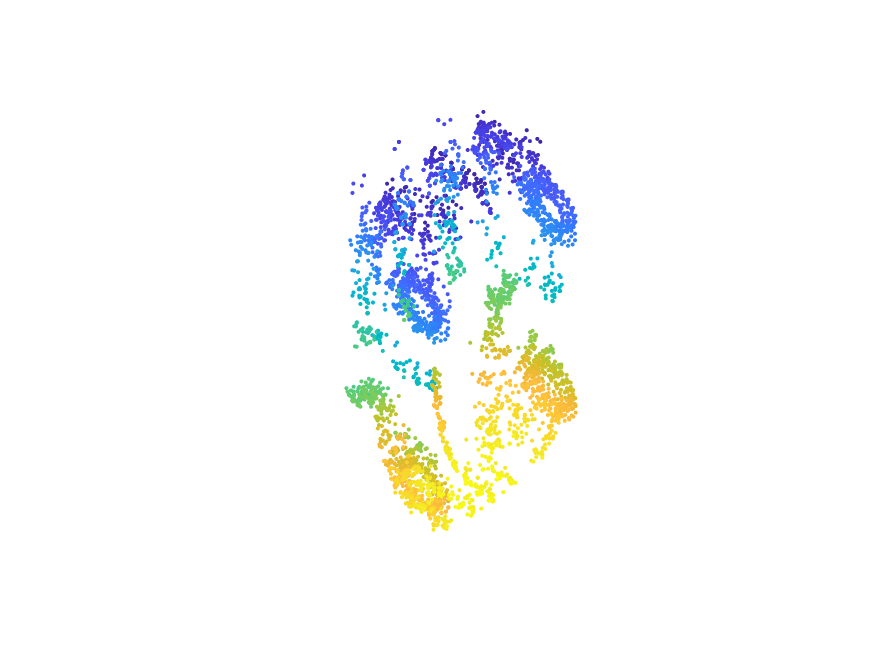}
\end{minipage}
\end{tabular}

\vspace{-1mm}
\caption{Demonstration of the 3D car model from~\cite{lin2014jointly} (in the form of point cloud) as the ground-truth shape for robust category-level perception.}
\label{demo-CR}
\vspace{-1mm}
\end{figure}

We can observe from Fig.~\ref{Benchmarking-PCR} that: (1) our TIVM is robust against 90\% outliers, in line with the 3 GNC solvers and ADAPT, and outperforming FGR (that fails at 70\%) and RANSAC \& FLO-RANSAC (that fail at 90\%), and (2) TIVM generally converges in merely 2--10 iterations and is the fastest general-purpose robust solver.
 
\textbf{Unknown Noise-Statistics Tests.} In Fig.~\ref{Benchmarking-PCR-NS-free}, we test the noise-statistics-free cases, where we find that: TIVM$^\circ$ still retains robustness and sufficient accuracy at 90\%  (rotation errors lower than 3$^\circ$ and translation errors lower than 0.02$m$) and also converges in about 10 iterations, up to 1 order of magnitude faster than the other 2 minimally-tuned solvers. 

\textbf{Real Application: Scan Stitching.} To validate our solver in real-world applications, we further test TIVM along with other competitors over the scan stitching problems using the 3DMatch dataset~\cite{zeng20173dmatch}. We use the FPFH descriptor~\cite{rusu2009fast} to establish correspondences between RGB-D scans and only preserve the test cases with outlier ratios lower than 90\% (because all general-purpose solvers fail at over 90\%) and $N>200$ (for ensuring sufficient correspondence number). The quantitative results in both known and unknown inlier-statistics scenarios are displayed in Fig.~\ref{3DMatch-PCR}, and some qualitative examples are illustrated in Fig.~\ref{3DMatch-PCR-qual}. We can see that our solver shows the best outlier-robustness and the highest time-efficiency ($\leq15$ iterations) in both scenarios.

\subsection{Category-Level Perception}

Robust category-level perception estimates both the pose: $\left(\boldsymbol{R}^{\star}\in SO(3),\boldsymbol{t}^{\star}\in \mathbb{R}^3\right)$ and the shape model: $\boldsymbol{c}^{\star}=[c^{\star}_1,c^{\star}_2,\dots,c^{\star}_K]^{\top}$ of the 3D object from correspondences between points $\{\boldsymbol{y}_i\}_{i=1}^N$ on the object and points on the given CAD models: $\{\mathcal{B}_k\}_{k=1}^K=\left\{\{\boldsymbol{b}_k(i)\}_{i=1}^N\right\}_{k=1}^K$, which are contaminated by outliers. (Readers can find more explicit problem definition in~\cite{shi2021optimal}.) These correspondences are described as: $\left\{\boldsymbol{y}_i\leftrightarrow\{\boldsymbol{b}_k(i)\}_{k=1}^K\right\}_{i=1}^N$, and the residual function can be represented as: $Re_i=\|\boldsymbol{R}^{\star}\sum^{K}_{k=1}c^{\star}_k\boldsymbol{b}_k(i)+\boldsymbol{t}^{\star}-\boldsymbol{y}_i\|_2$. 

\textbf{Setup.} We make use of the car models from the FG3DCar dataset~\cite{lin2014jointly} as the shape model, where we can find $K=15$ available CAD models over $N=256$ 3D points. In each run, we generate a random ground-truth pose: $\left(\boldsymbol{R}_{gt}\in SO(3),\boldsymbol{t}_{gt}\in \mathbb{R}^3\right)$ and a ground-truth shape model: $\boldsymbol{c}_{gt}$ to transform these car models and then add Gaussian noise with standard devation $\sigma=0.01$ on them to obtain the object points: $\{\boldsymbol{y}_i\}_{i=1}^N$. We replace some of the object points with random 3D points to produce outliers. The inlier threshold is set to $\tau=5\sigma$ for known noise-statistics tests. We apply the certifiably optimal method~\cite{shi2021optimal} as the non-minimal solver. Fig.~\ref{Benchmarking-CR} and~\ref{Benchmarking-CR-NS-free} show the results over 30 Monte Carlo runs in 2 noise-statistics scenarios, and the entire shape model in this experiment is displayed in Fig.~\ref{demo-CR}. 

TIVM remains robust at 80\% outliers, while all the other general-purpose non-minimal solvers yields wrong results at 80\%. The iteration number of TIVM is within 3--15 at all time, which is much more efficient than the other competitors. Although GNC-IRLS can converge in 8--20 iterations with $\leq$80\% outliers, it gets much slower at 90\%.

\textbf{Unknown Noise-Statistics Tests.} In Fig.~\ref{Benchmarking-CR-NS-free} where the noise-statistics-free tests are performed, we can see that the robustness of TIVM$^\circ$ is still 80\% and its iteration number till convergence is still within 15. ADAPT-MinT seems to show a lower estimation error at 80\% outliers, but it is 1--2 orders of magnitude less efficient than TIVM$^\circ$ and is less accurate at the lower-outlier regime.

\section{Conclusion}

In this paper, a general-purpose non-minimal robust estimator TIVM for robust geometric perception problems is rendered. By incorporating the multi-layered intra-class variance maximization technique with a self-adaptive layer-number tuning strategy into an iterative optimizing framework, an effective thresholding method for robust estimation by separating inliers from outliers is proposed. The special thresholding \& grouping mechanism of this estimator allows its robust performance even in the noise-statistics-free situations. Through validation experiments on various perception problems, the proposed estimator is shown to tolerate 70--90\% outliers with high accuracy no matter the inlier-noise statistics are given or not and usually require only 3--15 iterations to reach convergence, more efficient than the other general-purpose state-of-the-art robust solvers. Our demo code is provided at: \url{https://github.com/LeiSun-98/TIVM-master}.


{\small
\bibliographystyle{IEEEtran}
\bibliography{egbib}
}

\end{document}